\documentclass{ecai} 

\usepackage{latexsym}
\usepackage{amssymb}
\usepackage{amsmath}
\usepackage{amsthm}
\usepackage{booktabs}
\usepackage{enumitem}
\usepackage{graphicx}
\usepackage{color}

\usepackage{times}
\usepackage{soul}
\usepackage{url}
\usepackage[hidelinks]{hyperref}
\usepackage[utf8]{inputenc}
\usepackage[small]{caption}
\usepackage{graphicx}
\usepackage{booktabs}
\usepackage{algorithm}
\usepackage{algorithmic}
\usepackage[switch]{lineno}

\newtheorem{theorem}{Theorem}
\newtheorem{remark}{Remark}

\newtheorem{assumption}{Assumption}
\newtheorem{theorem_lemma}{Lemma}

\usepackage{colortbl}
\usepackage{microtype}
\usepackage{graphicx}
\usepackage{multirow}
\usepackage{mathtools}
\usepackage{subfigure}

\DeclareMathOperator*{\argmin}{arg\,min}
\newcommand*{\argminl}{\argmin\limits}
\usepackage{bm}
\usepackage{amsthm,amsmath,amssymb}
\usepackage{pifont}

\usepackage{wrapfig,lipsum,booktabs}

\newcommand{\BibTeX}{B\kern-.05em{\sc i\kern-.025em b}\kern-.08em\TeX}

\begin{document}


\begin{frontmatter}


\paperid{1809} 


\title{A First-Order Multi-Gradient Algorithm for Multi-Objective Bi-Level Optimization}



\author[A,B]{\fnms{Feiyang}~\snm{Ye}}
\author[C]{\fnms{Baijiong}~\snm{Lin}}
\author[D]{\fnms{Xiaofeng}~\snm{Cao}}
\author[A,E]{\fnms{Yu}~\snm{Zhang}\thanks{Corresponding Author. Email: yu.zhang.ust@gmail.com}}
\author[B,F,G,H]{\fnms{Ivor W.}~\snm{Tsang}}

\address[A]{Department of Computer Science and Engineering, Southern University of Science and Technology}
\address[B]{Australian Artificial Intelligence Institute, University of Technology Sydney}
\address[C]{The Hong Kong University of Science and Technology (Guangzhou)}
\address[D]{School of Artificial Intelligence, Jilin University}
\address[E]{Shanghai Artificial Intelligence Laboratory}
\address[F]{Centre for Frontier AI Research, Agency for Science, Technology and Research}
\address[G]{Institute of High Performance Computing, Agency for Science, Technology and Research}
\address[H]{School of Computer Science and Engineering, Nanyang Technological University}


\begin{abstract}
In this paper, we study the Multi-Objective Bi-Level Optimization (MOBLO) problem, where the upper-level subproblem is a multi-objective optimization problem and the lower-level subproblem is for scalar optimization. Existing gradient-based MOBLO algorithms need to compute the Hessian matrix, causing the computational inefficient problem. To address this, we propose an efficient first-order multi-gradient method for MOBLO, called FORUM. Specifically, we reformulate MOBLO problems as a constrained multi-objective optimization (MOO) problem via the value-function approach. Then we propose a novel multi-gradient aggregation method to solve the challenging constrained MOO problem. Theoretically, we provide the complexity analysis to show the efficiency of the proposed method and a non-asymptotic convergence result. Empirically, extensive experiments demonstrate the effectiveness and efficiency of the proposed FORUM method in different learning problems. In particular, it achieves state-of-the-art performance on three multi-task learning benchmark datasets. The code is available at \url{https://github.com/Baijiong-Lin/FORUM}.
\end{abstract}

\end{frontmatter}


\section{Introduction}\label{sec:introduction}

In this work, we study the {Multi-Objective Bi-Level Optimization} (MOBLO) problem, which is formulated as
\begin{equation}\label{eq:P1}
    \min_{\alpha\in \mathbb{R}^n, \omega \in \mathbb{R}^p} F(\alpha,\omega) \ ~~ \mathrm{s.t.} \ ~ \omega\in \mathcal{S}(\alpha) = \argminl_\omega f(\alpha,\omega),
\end{equation}
where $\alpha$ and $\omega$ denote the Upper-Level (UL) and Lower-Level (LL) variables, respectively. The UL subproblem, $F: = (F_1,F_2,\dots,F_m)^\top: \mathbb{R}^n \times \mathbb{R}^p \to \mathbb{R}^m$, is a vector-valued jointly continuous function for $m$ desired objectives. $\mathcal{S}(\alpha)$ denotes the optimal solution set (which is usually assumed to be a singleton set \cite{franceschi2017forward,ye2021multi}) of the LL subproblem by minimizing a continuous function $f(\alpha,\omega): \mathbb{R}^n \times \mathbb{R}^p \to \mathbb{R}$ w.r.t. $\omega$. In this work, we focus on MOBLO with a singleton set $\mathcal{S}(\alpha)$ and a non-convex UL subproblem, where $F_i$ is a non-convex function for all $i$. MOBLO has demonstrated its superiority in various learning problems such as neural architecture search \cite{liu2021multi,yue2020effective}, reinforcement learning \cite{abdolmaleki2020distributional,yang2019generalized}, multi-task learning \cite{ye2021multi}, and meta-learning \cite{ye2021multi,yuenhancing}.  

Recently, MOML \cite{ye2021multi} and MoCo \cite{fernando2023mitigating} are proposed as effective gradient-based MOBLO algorithms, which hierarchically optimize the UL and LL variables based on ITerative Differentiation (ITD) based Bi-Level Optimization (BLO) approach \cite{franceschi2017forward,franceschi2018bilevel,grazzi2020iteration}. Specifically, given $\alpha$, both MOML and MoCo first compute the LL solution $\omega^*(\alpha)$ by solving LL subproblem with $T$ iterations and then update $\alpha$ via the combination of the hypergradients $\{\nabla_{\alpha} F_i(\alpha,\omega^*(\alpha))\}_{i=1}^m$. Note that they need to calculate the complex gradient $\nabla_{\alpha} \omega^*(\alpha)$, which requires to compute many Hessian-vector products via the chain rule. Besides, their time and memory costs grow significantly fast with respect to the dimension of $\omega$ and $T$. Therefore, existing gradient-based methods to solve MOBLO problems could suffer from the inefficiency problem, especially in deep neural networks. 

\begin{table}[!t]
\centering
\caption{Comparison of convergence result and complexity analysis per UL iteration for different MOBLO methods. $m, n, p$, and $T$ denote the number of UL objectives, the dimension of the UL variables, the dimension of the LL variables, and the number of LL iterations, respectively.} 
\vskip 0.05in
\resizebox{\linewidth}{!}{\begin{tabular}{lcccc}
\toprule
\textbf{Method} & \textbf{Convergence analysis} & \textbf{Computational cost} & \textbf{Space cost}\\
\midrule
MOML \cite{ye2021multi} & asymptotic & $\mathcal{O}(mp(n+p)T)$ & $\mathcal{O}(mn+mpT)$ \\
MoCo \cite{fernando2023mitigating} & non-asymptotic  & $\mathcal{O}(mp(n+p)T)$ & $\mathcal{O}(2mn+mpT)$ \\
FORUM  & non-asymptotic  & $\mathcal{O}(mn+p(m+T))$ & $\mathcal{O}(mn+mp)$ \\
\bottomrule
\end{tabular}}
\label{tbl:add_table_compare}
\end{table}

To address this limitation, we propose an efficient \textbf{F}irst-\textbf{O}rde\textbf{R} m\textbf{U}lti-gradient method for \textbf{M}OBLO (\textbf{FORUM}). Specifically, we reformulate MOBLO as an equivalent constrained multi-objective optimization (MOO) problem by the value-function-based approach \cite{liu2021value,liubome,sow2022constrained}. Then, we propose a multi-gradient aggregation method to solve the challenging constrained MOO problem. Different from existing MOBLO methods such as MOML and MoCo, FORUM is a fully first-order algorithm and does not need to calculate the high-order Hessian matrix. Theoretically, we provide the complexity analysis showing that FORUM is more efficient than MOML and MoCo in both time and memory costs, as summarised in Table \ref{tbl:add_table_compare}. In addition, we provide a non-asymptotic convergence analysis for FORUM. Empirically, we evaluate the effectiveness and efficiency of FORUM on two learning problems, i.e., multi-objective data hyper-cleaning and multi-task learning on three benchmark datasets.

The main contributions of this work are three-fold: 
\begin{itemize}
\item We propose the FORUM method, an efficient gradient-based algorithm for the MOBLO problem; 
\item We demonstrate that the proposed FORUM method is more efficient than existing MOBLO methods from the perspective of complexity analysis and provide a non-asymptotic convergence analysis;
\item Extensive experiments demonstrate the effectiveness and efficiency of the proposed FORUM method. In particular, it achieves state-of-the-art performance on three benchmark datasets under the setting of multi-task learning.
\end{itemize}

\section{Related Works}\label{sec:relatedwork}
\paragraph{Multi-Objective Optimization.} 
MOO aims to solve multiple objectives simultaneously and its goal is to find the Pareto-optimal solutions. MOO algorithms can be broadly divided into three categories: population-based \cite{angus2007crowding}, evolutionary-based \cite{zhou2011multiobjective,chen2023evolutionary}, and gradient-based \cite{desideri12,mahapatra2020multi}. In this paper, we focus on the gradient-based category. One notable gradient-based MOO method is the MGDA \cite{desideri12} algorithm, which serves as a representative approach in this field. The MGDA algorithm employs a quadratic programming problem to determine the optimal direction for gradient updates during each training iteration. By doing so, it ensures that all objectives decrease simultaneously. Compared with the widely-used linear scalarization approach which linearly combines multiple objectives to a single objective, MGDA and its variants \cite{fernando2023mitigating,zhou2022convergence} have shown their superiority in many learning problems such as multi-task learning \cite{sk18} and reinforcement learning \cite{yu2020gradient}, especially when some objectives are conflicting.

\paragraph{Bi-Level Optimization.} BLO \cite{liu2021investigating} is a type of optimization problem with a hierarchical structure, where one subproblem is nested within another subproblem. The MOBLO problem (\ref{eq:P1}) reduces degrades to the BLO problem when $m$ equals $1$. One representative category of the BLO method is the ITD-based methods \cite{franceschi2017forward,franceschi2018bilevel,grazzi2020iteration} that use approximated hypergradient to optimize the UL variable, which is computed by the automatic differentiation based on the optimization trajectory of the LL variable. Some value-function-based algorithms \cite{liubome,liu2021value,sow2022constrained} have been proposed recently to solve BLO by reformulating the original BLO to an equivalent optimization problem with a simpler structure. The value-function-based reformulation strategy naturally yields a first-order algorithm, hence it has high computational efficiency. 

\paragraph{Multi-Objective Bi-Level Optimization.} 
MOML \cite{ye2021multi} is proposed as the first gradient-based MOBLO algorithm. However, MOML needs to calculate the complex Hessian matrix to obtain the hypergradient, causing the computationally inefficient problem. MoCo \cite{fernando2023mitigating} also employs the ITD-based approach like MOML for hypergradient calculation. It uses a momentum-like gradient approximation approach for hypergradient and a one-step approximation method to update the weights. It has the same inefficiency problem as the MOML method. \cite{yuenhancing} propose a mini-batch approach to optimize the UL subproblem in the MOBLO. However, it aims to generate weights for a huge number of UL objectives and is different from what we focus on. MORBiT \cite{gumin} studies a BLO problem with multiple objectives in its UL subproblem but it formulates the UL subproblem as a min-max problem, which is different from problem (\ref{eq:P1}) we focus on in this paper. 

\section{The FORUM Algorithm}\label{sec:solver}
In this section, we introduce the proposed FORUM method. Firstly, we reformulate MOBLO as an equivalent constrained multi-objective problem via the value-function-based approach in Section \ref{sec:3.1}. Next, we provide a novel multi-gradient aggregation method to solve the constrained multi-objective problem in Section \ref{sec:3.2}.

\subsection{Reformulation of MOBLO} \label{sec:3.1}

Based on the value-function-based approach \cite{liu2021value,liubome,sow2022constrained,kwon2023fully}, we reformulate MOBLO problem (\ref{eq:P1}) as an equivalent single-level \textit{constrained multi-objective optimization}   problem:
\begin{equation}\label{eq:FOMOBO_1}
    \min_{\alpha\in \mathbb{R}^n, \omega \in \mathbb{R}^p} F(\alpha,\omega) \ ~~ \mathrm{s.t.} \ ~ f(\alpha,\omega) \le f^*(\alpha),
\end{equation}
where $f^*(\alpha) = \min_{\omega}f(\alpha,\omega)=f(\alpha,\omega^*(\alpha))$ is the \textit{value function}, which represents the lower bound of $f(\alpha, \omega)$ w.r.t. $\omega$. To simplify the notation, we define $z\equiv(\alpha,\omega) \in \mathbb{R}^{n+p}$ and $\mathcal{Z}\equiv \mathbb{R}^n \times \mathbb{R}^p$. Then, we have $F(z)\equiv F(\alpha,\omega)$ and $f(z)\equiv f(\alpha,\omega)$. Thus, problem (\ref{eq:FOMOBO_1}) can be rewritten as
\begin{equation}\label{eq:FOMOBO_2}
\min_{z\in \mathcal{Z}} F(z) \ ~~ \mathrm{s.t.} \ ~ q(z) \le 0,
\end{equation}
where $q(z)= f(z)- f^*(\alpha)$ is the \textit{constraint function}. Since the gradient of the value function $f^*(\alpha)$ is 
\begin{equation}
    \nabla_{\alpha} f^*(\alpha) = \nabla_{\alpha} f(\alpha,\omega^*(\alpha)) = \nabla_{\alpha} f(\alpha,\omega^*),
\end{equation}
where the second equality is due to the chain rule and $\nabla_{\omega}f(\alpha,\omega)\mid_{\omega = \omega^*(\alpha)} = 0$, we do not need to compute the complex Hessian matrix $\nabla_{\alpha} \omega^*(\alpha)$ like MOML and MoCo. 

However, solving problem (\ref{eq:FOMOBO_2}) is challenging for two reasons. One reason is that the Slater's condition \cite{chen2023bilevel}, which is required for duality-based optimization methods, does not hold for problem (\ref{eq:FOMOBO_2}),  since the constraint $q(z)\le 0$ is ill-posed \cite{liu2021value,jiang2023conditional} and does not have an interior point. To see this, we assume $z_0 = (\alpha_0,\omega_0) \in \mathcal{Z}$ and $q(z_0)\le 0$. Then the constraint $q(z)\le 0$ is hard to be satisfied at the neighborhood of $\alpha_0$, unless $f^*(\alpha)$ is a constant function around $\alpha_0$, which rarely happens. Therefore, problem (\ref{eq:FOMOBO_2}) cannot be treated as classic constrained optimization and we propose a novel gradient method to solve it in Section \ref{sec:3.2}. Another reason is that for given $\alpha$, the computation of $\omega^*(\alpha)$ is intractable. Thus, we approximate it by $\tilde{\omega}^T$ computed by $T$ steps of gradient descent. 
Specifically, given $\alpha$ and an initialization $\tilde{\omega}^0$ of $\omega$, we have
\begin{equation}\label{op:5}
\tilde{\omega}^{t+1} = \tilde{\omega}^{t} - \eta \nabla_{\omega}f(\alpha, \tilde{\omega}^{t}),\ ~~ t=0,\cdots,T-1,
\end{equation}
where $\eta$ represents the step size. Then, the constraint function $q(z)$ is approximated by $\widetilde{q}(z)=f(z) - f(\alpha,\tilde{\omega}^T)$ and its gradient $\nabla_{z}q(z)$ is approximated by $\nabla_{z} \widetilde{q}(z)$. We show that the approximation error of the gradient $\nabla_{z} \widetilde{q}(z)$ exponentially decays w.r.t. the LL iterations $T$ in Appendix \ref{sec:add1}. Hence, problem (\ref{eq:FOMOBO_2}) is modified to
\begin{equation}\label{eq:FOMOBO_3}
\min_{z\in \mathcal{Z}} F(z) \ ~~ \mathrm{s.t.} \ ~ \widetilde{q}(z)= f(z) - f(\alpha,\tilde{\omega}^T) \le 0.
\end{equation}

\subsection{Multi-Gradient Aggregation Method}\label{sec:3.2}

In this section, we introduce the proposed multi-gradient aggregation method for solving problem (\ref{eq:FOMOBO_3}) iteratively. Specifically, at $k$-th iteration, assume $z_k$ is updated by $z_{k+1} = z_{k}+\mu d_k$ where $\mu$ is the step size and $d_k$ is the update direction for $z_k$. Then, we expect $d_k$ can simultaneously minimize the UL objective $F(z)$ and the constraint function $\widetilde{q}(z)$. Note that the minimum of the approximated constraint function $\widetilde{q}(z)$ converges to the minimum of $q(z)$, i.e. 0, as $T\to +\infty$. Thus, we expect $d_k$ to decrease $\widetilde{q}(z)$ consistently such that the constraint $\widetilde{q}(z)\le 0$ is satisfied.

Note that there are multiple potentially conflicting objectives $\{F_i\}_{i=1}^m$ in the UL subproblem. Hence, we expect $d_k$ can decrease every objective $F_i$, which can be formulated as the following problem to find $d_k$ to maximize the minimum decrease across all objectives as
\begin{equation}\label{eq:mgda}
\begin{split}
    \max_d\min_{i\in\{1,\dots,m\}}(&F_i(z_k) -  F_i(z_k+\mu d))\\
    &\approx -\mu \min_d\max_{i\in \{1,\dots,m\}} \langle \nabla F_i(z_k),d\rangle. 
\end{split}
\end{equation}
To regularize the update direction, we add a regularization term $\frac{1}{2}\|d\|^2$ to problem (\ref{eq:mgda}) and compute $d_k$ by solving 
$ \min_d\max_{i\in \{1,\dots,m\}} \langle \nabla F_i(z_k),d\rangle + \frac{1}{2}\|d\|^2$.

To decrease the constraint function $\widetilde{q}(z)$, we expect the inner product of $-d$ and $\nabla \widetilde{q}(z_k)$ to hold positive during the optimization process, i.e., $\langle \nabla \widetilde{q}(z_k),-d\rangle \ge \phi$, where $\phi$ is a non-negative constant. 

To further guarantee that $\widetilde{q}(z)$ can be optimized such that the constraint $\widetilde{q}(z)\le0$ can be satisfied, we introduce a dynamic $\phi_k$ here. Specifically, inspired by \cite{gong2021bi}, we set $\phi_k=\frac{\rho}{2}\|\nabla \widetilde{q}(z_k)\|^2$, where $\rho$ is a positive constant. When $\phi_k>0$, it means that $\|\nabla \widetilde{q}(z)\|\not=0$ and $\widetilde{q}(z)$ should be further optimized, and $\langle \nabla \widetilde{q}(z_k),-d\rangle \ge \phi_k > 0$ can enforce $\widetilde{q}(z)$ to decrease. When $\phi_k$ equals $0$, it indicates that the optimum of $\widetilde{q}(z)$ is reached and $\langle \nabla \widetilde{q}(z_k),-d\rangle \ge \phi_k = 0$ also holds. Thus, the dynamic $\phi_k$ can ensure $d_k$ to iteratively decrease $\widetilde{q}(z)$ such that the constraint $\widetilde{q}(z)\le0$ is satisfied.

Therefore, at $k$-th iteration, we can find $d_k$ by solving the problem:
\begin{equation}
\label{eq:modify_mgda}
\begin{split}
&\min_d\max_{i\in \{1,\dots,m\}} \langle \nabla F_i(z_k),d\rangle +\frac{1}{2}\|d\|^2, \\
&\ ~~~~~~ \mathrm{s.t.}\ \langle \nabla \widetilde{q}(z_k),d\rangle \le -\phi_k.
\end{split}
\end{equation}
Based on the Lagrangian multiplier method, problem (\ref{eq:modify_mgda}) has a solution as 
\begin{equation}\label{eq:dk}
d_k = -\left(\sum\nolimits_{i=1}^m \lambda_i^k\nabla F_i(z_k) + \nu(\lambda^k)\nabla \widetilde{q}(z_k)\right),
\end{equation}
where Lagrangian multipliers $\lambda^k=(\lambda^k_1,\ldots,\lambda^k_m)\in\Delta^{m-1}$ (i.e., $\sum_{i=1}^m \lambda_i^k=1$ and $\lambda_i^k\ge0$) and $\nu(\lambda)$ is a function of $\lambda$ as
\begin{equation}
\label{eq:FOMGDA_nu}
\begin{split}
    &\nu(\lambda) = \max\left(\sum\nolimits_{i=1}^m \lambda_i\pi_i(z_k), 0\right), \\
    &\ ~\mathrm{with} \ ~  \pi_i(z_k) = \frac{2\phi_k-\langle \nabla \widetilde{q}(z_k),\nabla F_i(z_k) \rangle}{\|\nabla \widetilde{q}(z_k)\|^2}.
\end{split}
\end{equation}
Here $\lambda_i^k$ can be obtained by solving the following dual problem as
\begin{equation}\label{eq:FOMGDA_main_text}
 \lambda^k = \argmin_{\lambda\in\Delta^{m-1}}\frac{1}{2}\left\|\sum\nolimits_{i=1}^m\lambda_i\nabla F_i(z_k) + \nu(\lambda)\nabla \widetilde{q}(z_k)\right\|^2 - \nu(\lambda)\phi_k.
\end{equation}
The detailed derivations of the above procedure are put in Appendix \ref{sec:add2}. Problem (\ref{eq:FOMGDA_main_text}) can be reformulated as
\begin{equation}
\label{eq:FOMGDA_2}
\begin{split}
&\min_{\lambda\in\Delta^{m-1},\gamma} \frac{1}{2}\left\|\sum\nolimits_{i=1}^m\lambda_i\nabla F_i(z_k) + \gamma\nabla \widetilde{q}(z_k)\right\|^2- \gamma\phi_k \\
&\ ~~~~~~ \mathrm{s.t.}\ \gamma\ge 0, \gamma\ge \sum\nolimits_{i=1}^m \lambda_i\pi_i(z_k).
\end{split}
\end{equation}
The first term of the objective function in problem (\ref{eq:FOMGDA_2}) can be simplified to $R^\top \Lambda^\top \Lambda R$, where $R = (\lambda_1, \ldots , \lambda_m, \gamma)^\top$ and $\Lambda = (\nabla F_1, \ldots ,\nabla F_m,\nabla \widetilde{q})$. Note that the dimension of the matrix $\Lambda^\top \Lambda$ is $(m+1)\times(m+1)$, which is independent with the dimension of $z$. As the number of UL objectives $m$ is usually small compared with the dimension of $z$, solving problem (\ref{eq:FOMGDA_2}) does not incur too much computational cost. In practice, we can use the open-source \texttt{CVXPY} library \cite{db16} to solve problem (\ref{eq:FOMGDA_2}).

To ensure convergence, the sequence of $\{\lambda^k\}_{k=1}^K$ should be a convergent sequence (refer to the discussion in Appendix \ref{sec:add3}). However, $\{\lambda^k\}_{k=1}^K$ obtained by directly solving the problem (\ref{eq:FOMGDA_2}) in each iteration cannot ensure such properties. Therefore, we apply a momentum strategy \cite{zhou2022convergence, yeadaptive} to $\lambda$ to generate a stable sequence and further guarantee the convergence. Specifically, in $k$-th iteration, we first solve the problem (\ref{eq:FOMGDA_2}) to obtain ${\lambda}^k$, then update the weights by 
$\tilde{\lambda}^k = (1-\beta_k) \tilde{\lambda}^{k-1}+ \beta_k{\lambda}^k$, 
where $\beta_k\in (0, 1]$ is set to $1$ at the beginning and asymptotically convergent to $0$ as $k\to +\infty$.

After obtaining $\tilde{\lambda}^k$ with the momentum update in $k$-th iteration, we can compute the corresponding $\nu(\tilde{\lambda}^k)$ via Eq. (\ref{eq:FOMGDA_nu}). Then we obtain the update direction $d_k$ by Eq. (\ref{eq:dk}) and update the variable $z_k$ as $z_{k+1}=z_k + \mu d_k$. The entire FORUM algorithm is shown in Algorithm \ref{alg:example}.  

\begin{algorithm}[!t]
\caption{The FORUM Method}
\label{alg:example}
\begin{algorithmic}[1]
\REQUIRE number of iterations $(K, T)$, step size $(\mu, \eta)$, coefficient $\beta_k$, constant $\rho$
\STATE Randomly initialize $z_0 = (\alpha_0,\omega_0)$;
\STATE Initialize $\tilde{\lambda}_i^{-1}=1/m$, $i=1,...,m$;
\FOR{$k=0$ {\bfseries to} $K-1$}
\STATE Set $\tilde{\omega}^0 = \omega_0$ or $\tilde{\omega}^0 = \omega_k$;
\FOR{$t=0$ {\bfseries to} $T-1$}
\STATE Update $\tilde{\omega}$ as $\tilde{\omega}^{t+1} = \tilde{\omega}^{t} - \eta \nabla_{\omega}f(\alpha_k, \tilde{\omega}^{t})$;
\ENDFOR
\STATE Set $\widetilde{q}(z_k) = f(z_k) -f(\alpha_k,\tilde{\omega}^{T}) $;
\STATE Compute gradient $\nabla_z \widetilde{q}(z_k) = \nabla_z f(z_k) -\nabla_{\alpha}f(\alpha_k,\tilde{\omega}^{T}) $;
\STATE Compute gradients $\nabla_{z} F_i(z_k),~i=1,\dots,m$;
\STATE Compute ${\lambda}^k$ by solving problem (\ref{eq:FOMGDA_2});
\STATE Update $\tilde{\lambda}^k$ by $\tilde{\lambda}^k = (1-\beta_k) \tilde{\lambda}^{k-1}+ \beta_k{\lambda}^k$;
\STATE Compute $\nu(\tilde{\lambda}^k)$ via Eq. (\ref{eq:FOMGDA_nu});
\STATE Compute $d_k$ via Eq. (\ref{eq:dk});
\STATE Update $z$ as $z_{k+1} = z_k +\mu d_k$;
\ENDFOR
\STATE {\textbf{return}} $z_K$.
\end{algorithmic}
\end{algorithm}
\vskip 0.1in

\section{Analysis}
\label{sec:Analysis}

In this section, we provide complexity analysis and convergence analysis for the FORUM method.

\subsection{Complexity Analysis} \label{sec:com_analysis}

For the proposed FORUM method, it takes time $\mathcal{O}(pT)$ and space $\mathcal{O}(p)$ to obtain the approximated constraint function $\widetilde{q}(z)$, and then the computations of all the gradients including $\nabla_z F_i(z)$ and $\nabla_z \widetilde{q}(z)$ require time $\mathcal{O}((n+p)(m+1))$ and space $\mathcal{O}((n+p)(m+1))$. 
When the number of UL objectives $m$ satisfies $m\ll n+p$, the time and space costs of solving the quadratic programming problem (\ref{eq:FOMGDA_2}), which only depends on $m$, can be negligible.
Therefore, FORUM runs in time $\mathcal{O}(mn+p(m+T))$ and space $\mathcal{O}(mn+mp)$ in total for each UL iteration. 

We provide a complexity analysis for the existing MOBLO methods (i.e., MOML and MoCo). For the MOML method, it takes $\mathcal{O}(pT)$ time and $\mathcal{O}(p)$ space to do the $T$-iteration update for the LL subproblem. Then calculating the Hessian-matrix product via backward propagation in each UL iteration can be evaluated in time $\mathcal{O}(p(n+p)T)$ and space $\mathcal{O}(n+pT)$. Similar to the FORUM method, the cost of solving the quadratic programming problem in MOML is also negligible. Therefore, for each UL iteration, MOML require $\mathcal{O}(mp(n+p)T)$ time and  $\mathcal{O}(mn+mpT)$ space in total. For the MoCo method, it uses a similar approach to MOML to calculate the Hessian-matrix product via backward propagation in each UL iteration. Note that MoCo applies a momentum update to the UL variables $\alpha$, which causes an additional $\mathcal{O}(mn)$ space cost. Thus, for each UL iteration, MoCo require $\mathcal{O}(mp(n+p)T)$ time and $\mathcal{O}(2mn+mpT)$ space in total.

In summary, the above analysis indicates that the proposed FORUM method is more efficient than MOML and MoCo in terms of both time and space complexity.

\subsection{Convergence Analysis} \label{sec:con_analysis}
In this section, we analyze the convergence property of FORUM. Firstly, we make an assumption for the UL subproblem.
\begin{assumption}\label{assume:1}
For $i=1,\dots,m$, it is assumed that the gradient $\nabla F_i(\alpha,\omega)$ is $L_F$-Lipschitz continuous with respect to $z:= (\alpha,\omega)$. The $\ell_2$ norm of $\nabla F_i(z)$ and $|F_i(z)|$ are upper-bounded by a positive constant $M$.
\end{assumption}
The smoothness and the boundedness assumptions in Assumption \ref{assume:1} are widely adopted in non-convex multi-objective optimization \cite{zhou2022convergence,fernando2023mitigating}. Then we make an assumption for the LL subproblem.
\begin{assumption}\label{assume:2}
The function $f(\alpha,\omega)$ is $c$-strongly convex with respect to $\omega$, and the gradient $\nabla f(\alpha,\omega)$ is $L_f$-Lipschitz continuous with respect to $z:= (\alpha,\omega)$.
\end{assumption}
The strongly convexity assumption in Assumption \ref{assume:2} is commonly used in the analysis for the BLO \cite{franceschi2017forward,franceschi2018bilevel} and MOBLO problems \cite{fernando2023mitigating,ye2021multi}. The proposed FORUM algorithm focuses on generating one Karush-Kuhn-Tucker (KKT) stationary point of the original constrained multi-objective optimization problem (\ref{eq:FOMOBO_2}). Following \cite{gong2021bi,liubome}, we measure the convergence of problem (\ref{eq:FOMOBO_2}) by both its KKT stationary condition and the feasibility condition, where detailed definitions are provided in Appendix \ref{sec:app_analysis_1}. Specifically, we denote by $\mathcal{K}(z_k) = \left\|\sum_{i=1}^{m}\tilde{\lambda}^k_i \nabla F_i(z_k) + \nu_k \nabla q(z_k)\right\|^2$ the measure of KKT stationary condition in the $k$-th iteration, where $\nu_k=\nu(\tilde{\lambda}^k)$. To satisfy the feasibility condition of problem (\ref{eq:FOMOBO_2}), the non-negative function $q(z_k)$ should decrease to $0$. Then, with a non-convex multi-objective UL subproblem, we have the following convergence result.
\begin{theorem}\label{thm:2}
Suppose that Assumptions \ref{assume:1} and \ref{assume:2} hold, and the sequence $\{z_k\}_{k=0}^K$ generated by Algorithm \ref{alg:example} satisifes $q(z_k)\le B$, where $B$ is a positive constant. Then if $\eta\le 1/L_f$, $\mu=\mathcal{O}(K^{-1/2})$ , and $\beta=\mathcal{O}(K^{-3/4})$, there exists a constant $C>0$ such that when $T \ge C$, for any $K>0$, we have
\begin{align}\label{eq:thm2}
\max\left\{\min_{k<K}\mathcal{K}(z_k),q(z_k)\right\} = 
       \mathcal{O}(K^{-1/4}+\Gamma(T)),
   \end{align}
   where $\Gamma(T)$ represents exponential decays with respect to $T$.
\end{theorem}

The proof is put in Appendix \ref{sec:app_analysis_3}. Theorem \ref{thm:2} gives a non-asymptotic convergence result for Algorithm \ref{alg:example} based on the KKT stationary condition and the feasibility condition of the problem (\ref{eq:FOMOBO_2}). The proposed FORUM method achieves a $\mathcal{O}(K^{-1/4}+\Gamma(T))$ convergent rate, which depends on both numbers of steps in the UL and LL subproblems (i.e., $K$ and $T$). 

\begin{figure*}[!t]
\centering
\subfigure[Running time vs. $T$.]{\label{fig:a}\includegraphics[width=0.24\textwidth]{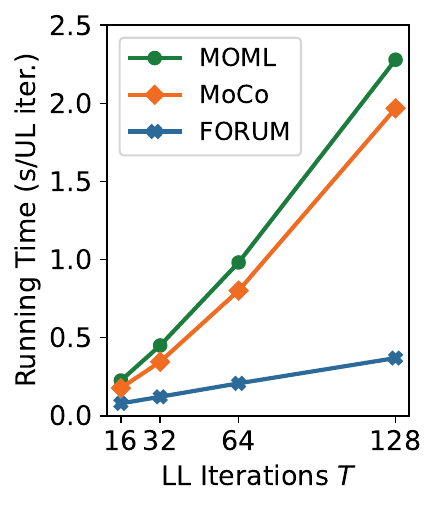}}
\hfill
\subfigure[Running time vs. $p$.]
{\label{fig:b}\includegraphics[width=0.235\textwidth]{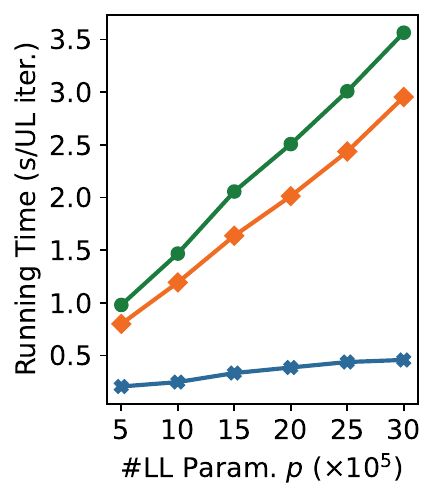}}
\hfill
\subfigure[Memory cost vs. $T$.]
{\label{fig:c}\includegraphics[width=0.24\textwidth]{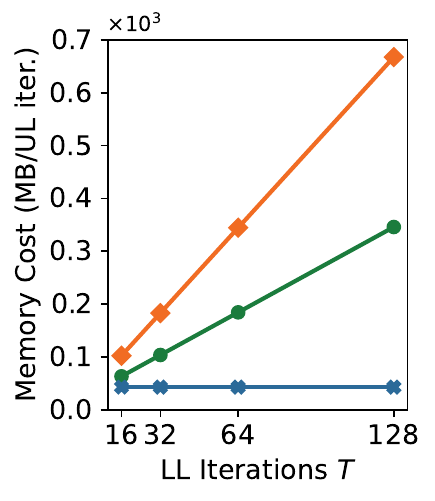}}
\hfill
\subfigure[Memory cost vs. $p$.]
{\label{fig:d}\includegraphics[width=0.24\textwidth]{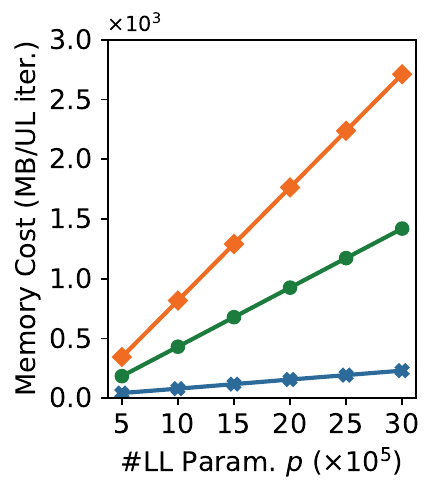}}
\caption{Results of different MOBLO methods on the multi-objective data hyper-cleaning problem. \textbf{(a)}: The running time per iteration varies over different LL update steps $T$ with fixed numbers of LL parameters $p$. \textbf{(b)}: The running time per iteration varies over the different numbers of LL parameters $p$ with $T=64$. \textbf{(c)}: The memory cost varies over different LL update steps $T$ with fixed numbers of LL parameters $p$.
\textbf{(d)}: The memory cost varies over the different numbers of LL parameters $p$ with $T=64$.  
}
\vskip 0.2in
\label{fig:1}
\end{figure*}

\section{Experiments} \label{sec:exp}

In this section, we empirically evaluate the proposed FORUM method on different learning problems. 
All experiments are conducted on a single NVIDIA GeForce RTX 3090 GPU. 

\subsection{Data Hyper-Cleaning}

\paragraph{Setup.}
Data hyper-cleaning \cite{bao2021stability,franceschi2017forward,liubome,shaban2019truncated} is a hyperparameter optimization problem, where a model is trained on a dataset with part of training labels corrupted. Thus, it aims to reduce the influence of noisy examples by adding weights
to the train samples and learning these weights in a bi-level optimization manner. Here we extend data hyper-cleaning to a multi-objective setting, where we aim to train a model on multiple corrupted datasets.

Specifically, suppose that there are $m$ corrupted datasets. $\mathcal{D}^{\text{tr}}_i= \{x_{i,j},y_{i,j}\}_{j=1}^{N_i}$ and $\mathcal{D}^{\text{val}}_i$ denote the noisy training set and the clean validation set for the $i$-th dataset, respectively, where $x_{i,j}$ denotes the $j$-th training sample in the $i$-th dataset, $y_{i,j}$ is the corresponding label, and $N_i$ denotes the size of the $i$-th training dataset. Let $\omega$ denote the model parameters and $\alpha_{i,j}$ denotes the weight of the training sample $x_{i,j}$.  Let $\mathcal{L}^{\text{val}}_i( \omega;\mathcal{D}^{\text{val}}_i)$ be the average loss of model $\omega$ on the clean validation set of the $i$-th dataset and 
\begin{equation*}
    \mathcal{L}^{\text{tr}}_i(\alpha, \omega;\mathcal{D}^{\text{tr}}_i) = \frac{1}{N_i}\sum\nolimits_{j=1}^{N_i}\sigma(\alpha_{i,j})\ell(\omega;x_{i,j},y_{i,j})
\end{equation*}
be the weighted average loss on the noisy training set of the $i$-th dataset, where $\sigma(\cdot) $ is an element-wise sigmoid function to constrain each weight in the range $[0,1]$ and $\ell(\omega;x,y)$ denotes the loss of model $\omega$ on sample $(x,y)$. Therefore, the objective function of this multi-objective data hyper-cleaning is formulated as
\begin{equation*}
\begin{split}
    &\min_{\alpha,\omega}\left(\mathcal{L}_1^{\text{val}}(\omega;\mathcal{D}^{\text{val}}_1),\cdots, \mathcal{L}_m^{\text{val}}(\omega;\mathcal{D}^{\text{val}}_m)\right)^\top\\
    &\ ~ \mathrm{s.t.}\ ~ \omega\in \mathcal{S}(\alpha)=\argminl_{\omega} \sum\nolimits_{i=1}^m {\mathcal{L}}_i^{\text{tr}}(\alpha, \omega;\mathcal{D}^{\text{tr}}_i).
\end{split}
\end{equation*}

\paragraph{Datasets.} We conduct experiments on the MNIST \cite{lecun1998gradient} and FashionMNIST \cite{xiao2017fashion} datasets. Each dataset corresponds to a $10$-class image classification problem. All the images have the same size of $28\times 28$. Following \cite{bao2021stability}, we randomly sample $5000$, $1000$, $1000$, and $5000$ images from each dataset as the training set, first validation set, second validation set, and test set, respectively. The training set and first validation set are used to formulate the LL and UL subproblems, respectively. The second validation set is used to select the best model and the testing evaluation is conducted on the test set. Half of the samples in the training set are contaminated by assigning them to another random class.

\paragraph{Implementation Details.} The proposed FORUM method is compared with 
two \textit{MOBLO methods}: MOML \cite{ye2021multi} and MoCo \cite{fernando2023mitigating}.
The same configuration is used for both the MOML, MoCo, and FORUM methods. 
Specifically, the hard-parameter sharing architecture \cite{zhang2021survey} is used, where the bottom layers are shared among all datasets and each dataset has its specific head layers. The shared module contains two linear layers with input size, hidden size, and output size of $784$, $512$, and $256$. Each layer adopts a ReLU activation function. Each dataset has a specific linear layer with an output size of $10$. The batch size is set to $100$.
For the LL subproblem, the SGD optimizer with a learning rate $\eta = 0.3$ is used for updating $T=64$ iterations. For the UL subproblem, the total number of UL iterations $K$ is set to $1200$, and an SGD optimizer with the learning rate as $10$ is used for updating weight $\alpha$ while another SGD optimizer with the learning rate as $0.3$ is used for updating model parameters $\omega$.
We set $\rho=0.5$ and $\beta_k=(k+1)^{-\frac{3}{4}}$ for FORUM.
For Figures \ref{fig:b} and \ref{fig:d}, we increase the number of LL parameters $p$ by adding some linear layers with the hidden size of $512$ into the shared module. 
We use the build-in function \textsf{torch.cuda.max\_memory\_allocated} in PyTorch \cite{paszke2019pytorch} to compute the GPU memory cost in Figures \ref{fig:c} and \ref{fig:d}.  

\begin{table}[!t]
\centering
\caption{Performance of different methods on the MNIST and FashionMNIST datasets for the multi-objective data hyper-cleaning problem. Each experiment is repeated over $3$ random seeds, and the mean and the standard deviation are reported. The best result is marked in \textbf{bold}.}
\vskip 0.1in
\resizebox{\linewidth}{!}{
\begin{tabular}{lcccc}
\toprule
\multirow{2.5}{*}{\textbf{Methods}} & \multicolumn{2}{c}{\textbf{MNIST}} & \multicolumn{2}{c}{\textbf{FashionMNIST}}\\
\cmidrule(lr){2-3} \cmidrule(l){4-5}
& \textbf{Accuracy (\%)} & \textbf{F1 Score} & \textbf{Accuracy (\%)} & \textbf{F1 Score}\\
\midrule
MOML \cite{ye2021multi} & $88.64_{\pm0.94}$ & $88.61_{\pm0.98}$ & $80.64_{\pm0.35}$ & $80.60_{\pm0.49}$ \\
MoCo \cite{fernando2023mitigating}  & $88.05_{\pm1.21}$ & $88.03_{\pm1.27}$ & $80.94_{\pm0.19}$ & $80.67_{\pm0.25}$\\
FORUM & $\textbf{90.81}_{\pm0.14}$ & $\textbf{90.81}_{\pm0.15}$ & $\textbf{82.07}_{\pm0.38}$ & $\textbf{81.72}_{\pm0.57}$\\
\bottomrule
\end{tabular}}
\vskip 0.1in
\label{tab:cleaning}
\end{table}

\paragraph{Results.} 
 The classification accuracy and F1 score computed on the test set are used as the evaluation metrics. The results are provided in Table \ref{tab:cleaning}. As can be seen, the proposed FORUM method outperforms the MOML and MoCo in both datasets, which demonstrates its effectiveness. 

Figures \ref{fig:a} and \ref{fig:b} show that MOML and MoCo need longer running time than FORUM in every configuration of the UL iteration $T$ and the number of LL parameters $p$, respectively, which implies FORUM has a lower time complexity. Figures \ref{fig:c} and \ref{fig:d} show the change of memory cost per iteration with respect to the LL iteration $T$ and the number of LL parameters $p$, respectively. As can be seen, the memory cost remains almost constant with different $T$'s for FORUM and increases faster for MOML and MoCo.  Moreover, the memory cost slightly increases in FORUM with increasing $p$, while it linearly increases in MOML and MoCo. In summary, the results in Figure \ref{fig:1} match the complexity analysis in Section \ref{sec:com_analysis} and demonstrate that FORUM is more efficient than MOML and MoCo in terms of both time and space complexity.

\begin{figure}[!t]
\centering
\subfigure[Accuracy vs. $\rho$.]
{\includegraphics[width=0.48\linewidth]{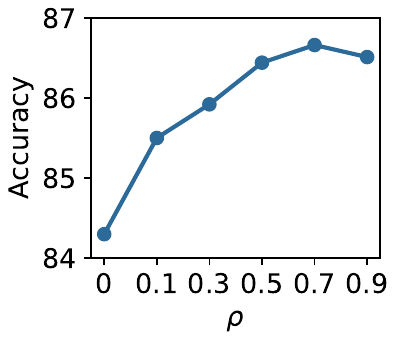}}
\hfill
\subfigure[Accuracy vs. $\eta$.]
{\includegraphics[width=0.48\linewidth]{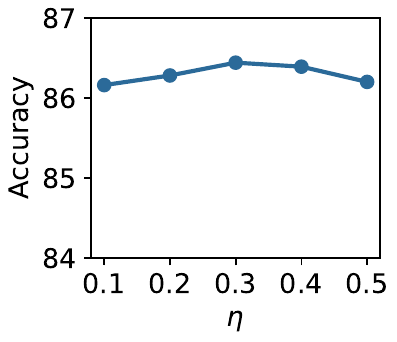}}
\caption{Effects of $\rho$ (\textbf{Left}) and $\eta$ (\textbf{Right}) in the multi-objective data hyper-cleaning problem. ``Accuracy'' denotes the average accuracy on MNIST and FashionMNIST datasets.}
\vskip 0.3in
\label{fig:rho_eta}
\end{figure}

\paragraph{Effects of $\eta$ and $\rho$.} We conduct additional experiments to study the effects of hyperparameters $\eta$ and $\rho$ in the data hyper-cleaning problem. The results are shown in Figure \ref{fig:rho_eta}. FORUM is insensitive with $\eta$ and a large $\rho$ (e.g, $\rho=0.5, 0.7, 0.9$). Besides, FORUM with a positive $\rho$ performs better than $\rho=0$, which shows the effectiveness of $\phi_k$ introduced in Section \ref{sec:3.2}.  

\subsection{Multi-Task Learning} \label{sec:mtl}

\paragraph{Setup.}
Multi-Task Learning (MTL) \cite{zhang2021survey, lin2023scale} aims to train a single model to solve several tasks simultaneously. Following MOML \cite{ye2021multi}, we aim to learn the loss weights to balance different tasks and improve the generalization performance by casting MTL as a MOBLO problem. 
Specifically, suppose there are $m$ tasks and the $i$-th task has its corresponding dataset $\mathcal{D}_i$ that contains a training set $\mathcal{D}_i^{\text{tr}}$ and a validation set $\mathcal{D}_i^{\text{val}}$. The MTL model is parameterized by $\omega$ and $\alpha \in \Delta^{m-1}$ denotes the loss weights for the $m$ tasks. Let $\mathcal{L}(\omega;\mathcal{D})$ represent the average loss of model $\omega$ on the dataset $\mathcal{D}$. The MOBLO formulation for MTL is as 
\begin{equation*}
\begin{split}
    &\min_{\alpha,\omega}~\big(\mathcal{L}(\omega;\mathcal{D}^{\text{val}}_1),\cdots,\mathcal{L}(\omega;\mathcal{D}^{\text{val}}_m)\big)^\top\\
    &\ ~ \mathrm{s.t.}\ ~ \omega\in \mathcal{S}(\alpha)=\argminl_{{\omega}} \sum\nolimits_{i=1}^m \alpha_i\mathcal{L}(\omega;\mathcal{D}^{\text{tr}}_i).
\end{split}
\end{equation*}
We conduct experiments on three benchmark datasets among three different task categories, i.e., the Office-31 \cite{saenko2010adapting} dataset for the image classification task, the NYUv2 \cite{silberman2012indoor} dataset for the scene understanding task, and the QM9 dataset \cite{ramakrishnan2014quantum} for the molecular property prediction task. 

\begin{table}[!t]
\centering
\caption{Classification accuracy (\%) on Office-31. Each experiment is repeated over 3 random seeds and the average is reported. The best results over baselines except STL are marked in \textbf{bold}. } 
\vskip 0.1in
\resizebox{0.95\linewidth}{!}{
\begin{tabular}{lccccc}
\toprule
\textbf{Methods} & 
{\textbf{A}} & {\textbf{D}} & {\textbf{W}} & {\textbf{Avg}} & \bm{$\Delta_{\mathrm{p}}$}${\uparrow}$\\
\midrule
STL & 86.61 & 95.63 & 96.85 & 93.03 & 0.00\\
\midrule
\multicolumn{6}{c}{\textit{multi-task learning methods}} \\
EW \cite{zhang2021survey} & 83.53 & 97.27 & 96.85 & 92.55 & -0.61\\
UW \cite{kgc18} & 83.82 & 97.27 & 96.67 & 92.58 & -0.56\\
MGDA \cite{sk18} & 85.47 & 95.90 & 97.03 & 92.80 & -0.27\\
PCGrad \cite{yu2020gradient}  & 83.59 & 96.99 & 96.85 & 92.48 & -0.68\\
GradDrop \cite{chen2020just} & 84.33 & 96.99 & 96.30 & 92.54 & -0.59\\
GradVac \cite{wang2021gradient} & 83.76 & 97.27 & 96.67 & 92.57 & -0.58\\
CAGrad \cite{liu2021conflict} & 83.65 & 95.63 & 96.85 & 92.04 & -1.13\\
IMTL \cite{liu2021towards} & 83.70 & 96.44 & 96.29& 92.14& -1.02\\
Nash-MTL \cite{navon2022multi} & 85.01 & 97.54 & 97.41 & 93.32 & +0.24\\
RLW \cite{linreasonable} & 83.82 & 96.99 & 96.85 & 92.55 & -0.59\\
\midrule
\multicolumn{6}{c}{\textit{first-order bi-level optimization methods}} \\
BVFIM \cite{liu2021value} & 84.84 & 96.99 & 97.78 & 93.21 & +0.11\\
BOME \cite{liubome} & 85.53 & 96.72 & \textbf{98.15} & 93.47 & +0.41\\
\midrule
\multicolumn{6}{c}{\textit{multi-objective bi-level optimization methods}} \\
MOML \cite{ye2021multi}  & 84.67 & 96.72 & 96.85 & 92.75 & -0.36\\
MoCo \cite{fernando2023mitigating} & 84.38 & 97.26 & 97.03 & 92.89 & -0.22\\
FORUM & \textbf{85.64} & \textbf{98.63} & 97.96 & \textbf{94.07} & \textbf{+0.96}\\
\bottomrule
\end{tabular}}
\vskip 0.1in
\label{tbl:mtl-office31}
\end{table}

\begin{table*}[!t]
\centering
\caption{Results on the {NYUv2} dataset. Each experiment is repeated over 3 random seeds and the average is reported. The best results over baselines except STL are marked in \textbf{bold}. $\uparrow$ ($\downarrow$) indicates that the higher (lower) the result, the better the performance.}
\vskip 0.1in
\resizebox{0.9\textwidth}{!}{
\begin{tabular}{lcccccccccc}
\toprule
\multirow{4}{*}{\textbf{Methods}} & \multicolumn{2}{c}{\textbf{Segmentation}} & \multicolumn{2}{c}{\textbf{Depth}} & \multicolumn{5}{c}{\textbf{Surface Normal Prediction}} & 
\multirow{4.5}{*}{\bm{$\Delta_{\mathrm{p}}$}${\uparrow}$}\\
\cmidrule(lr){2-3} \cmidrule(lr){4-5} \cmidrule(lr){6-10}
& \multirow{2.5}{*}{\textbf{mIoU${\uparrow}$}} &  \multirow{2.5}{*}{\textbf{PAcc$\uparrow$}} &  \multirow{2.5}{*}{\textbf{AErr$\downarrow$}} &  \multirow{2.5}{*}{\textbf{RErr$\downarrow$}} & \multicolumn{2}{c}{\textbf{Angle Distance}} & \multicolumn{3}{c}{\textbf{Within $t^{\circ}$}} \\ \cmidrule(lr){6-7} \cmidrule(lr){8-10} & & & & & \textbf{Mean$\downarrow$} & \textbf{Median$\downarrow$}  & \textbf{11.25$\uparrow$} & \textbf{22.5$\uparrow$} & \textbf{30$\uparrow$}  \\
\midrule
STL & 53.50 & 75.39 & 0.3926 & 0.1605 & 21.9896 & 15.1641 & 39.04 & 65.00 & 75.16 & 0.00 \\
\midrule
\multicolumn{11}{c}{\textit{multi-task learning methods}} \\
EW \cite{zhang2021survey} & 53.93 & 75.53 & 0.3825 & 0.1577 & 23.5691 & 17.0149 & 35.04 & 60.99 & 72.05 & -1.78\\
UW \cite{kgc18} & \textbf{54.29} & 75.64 & 0.3815 & 0.1583 & 23.4805 & 16.9206 & 35.26 & 61.17 & 72.21 & -1.52 \\
MGDA \cite{sk18} & 53.52 & 74.76 & 0.3852 & 0.1566 & 22.7400 & 16.0000 & 37.12 & 63.22 & 73.84 & -0.64\\
PCGrad \cite{yu2020gradient} & 53.94 & 75.62 & 0.3804 & 0.1578 & 23.5226 & 16.9276 & 35.19 & 61.17 & 72.19 & -1.57\\
GradDrop \cite{chen2020just} & 53.73 & 75.54 & 0.3837 & 0.1580 & 23.5392 & 16.9587 & 35.17 & 61.06 & 72.07 & -1.85\\
GradVac \cite{wang2021gradient} & {54.21} & 75.67 & 0.3859 & 0.1583 & 23.5804 & 16.9055 & 35.34 & 61.15 & 72.10 & -1.75\\
CAGrad \cite{liu2021conflict} & 53.97 & 75.54 & 0.3885 & 0.1588 & 22.4701 & 15.7139 & \textbf{37.77} & 63.82 & 74.30 & -0.27\\
IMTL \cite{liu2021towards} &53.63 &75.44& 0.3868 &0.1592& 22.5800 &15.8500&37.44 &63.52 &74.09 &-0.57\\
Nash-MTL \cite{navon2022multi} & 53.41 & 74.95 & 0.3867 & 0.1612 & 22.5662 & 15.9365 & 37.30 & 63.40 & 74.09 & -1.01\\
RLW \cite{linreasonable} & 54.13 & 75.72 & 0.3833 & 0.1590 & 23.2125 & 16.6166 & 35.88 & 61.84 & 72.74 & -1.27\\
\midrule
\multicolumn{11}{c}{\textit{first-order bi-level optimization methods}} \\
BVFIM \cite{liu2021value} & 53.29 & 75.07 & 0.3981 & 0.1632 & 22.3552 & 15.9710 & 37.15 & 63.44 & 74.27 & -1.68\\
BOME \cite{liubome} & 54.15 & \textbf{75.79} & 0.3831 & 0.1578 & 23.3378 & 16.8828 & 35.29 & 61.31 & 72.40 & -1.45\\
\midrule
\multicolumn{11}{c}{\textit{multi-objective bi-level optimization methods}} \\
MOML \cite{ye2021multi} & 53.59 & 75.48 & 0.3839 & 0.1577 & 23.1487 & 16.5319 & 36.06 & 62.05 & 72.89 & -1.26\\
MoCo \cite{fernando2023mitigating} & 53.73 & 75.63 & 0.3838 & 0.1560 & 23.1922 & 16.5737 & 36.02 & 61.93 & 72.82 & -1.06\\
FORUM & 54.04 & 75.64 & \textbf{0.3795} & \textbf{0.1555} & \textbf{22.1870} & \textbf{15.6815} & 37.71 & \textbf{64.04} & \textbf{74.67} & +\textbf{0.65}\\
\bottomrule
\end{tabular}}
\vskip 0.1in
\label{tbl:mtl-nyu_dmtl}
\end{table*}

\paragraph{Datasets.} 
The {Office-31} dataset \cite{saenko2010adapting} includes images from three different sources: Amazon ({A}), digital SLR cameras ({D}), and Webcam ({W}). It contains $31$ categories for each source and a total of $4652$ labeled images. We use the data split in RLW \cite{linreasonable}: $60\%$ for training, $20\%$ for validation, and $20\%$ for testing. 
The {NYUv2} dataset \cite{silberman2012indoor}, an indoor scene understanding dataset, has $795$ and $654$ images for training and testing, respectively. It has three tasks: $13$-class semantic segmentation, depth estimation, and surface normal prediction. 
The {QM9} dataset \cite{ramakrishnan2014quantum}, a molecular property prediction dataset. We use the same data split as in Nash-MTL \cite{navon2022multi}: $110, 000$ for training, $10,000$ for validation, and $10,000$ for testing. The {QM9} dataset contains $11$ tasks and each task is a regression task for one property. 

\paragraph{Baselines.} The proposed FORUM method is compared with a number of baseline methods from four different categories: 
\textit{single-task learning} (STL) method that trains each task independently; 
a comprehensive set of state-of-the-art \textit{MTL methods}, including Equal Weighting (EW) \cite{zhang2021survey}, UW \cite{kgc18}, MGDA \cite{sk18}, PCGrad \cite{yu2020gradient}, GradDrop \cite{chen2020just}, GradVac \cite{wang2021gradient}, CAGrad \cite{liu2021conflict}, IMTL\cite{liu2021towards}, Nash-MTL \cite{navon2022multi}, and RLW \cite{linreasonable}; 
two \textit{first-order BLO methods}: BVFIM \cite{liu2021value} and BOME \cite{liubome}, where we simply transform MOBLO to BLO by aggregating multiple objectives in the UL subproblem with equal weights into a single objective so that we can apply those BLO methods to solve the MOBLO problem; 
two \textit{gradient-based MOBLO methods}: MOML \cite{ye2021multi} and MoCo \cite{fernando2023mitigating}.

\paragraph{Evaluation Metrics.} 
For the Office-31 dataset, following RLW \cite{linreasonable}, we use classification accuracy as the evaluation metric for each task and the average accuracy as the overall metric. 
For the NYUv2 dataset, following RLW \cite{linreasonable}, we use the mean intersection over union (MIoU) and the class-wise pixel accuracy (PAcc) for the semantic segmentation task, the relative error (RErr) and the absolute error (AErr) for the depth estimation task, and the mean and median angle error as well as the percentage of normals within $t^{\circ}~(t=11.25, 22.5, 30)$ for the surface normal prediction task. 
For the {QM9} dataset, following Nash-MTL \cite{navon2022multi}, we use mean absolute error (MAE) as the evaluation metric. 

Following \cite{linreasonable,lin2023scale,ye2023multi}, we use $\Delta_{\mathrm{p}}$ as a metric to evaluate the overall performance on all the tasks. It is defined as the mean of the relative improvement of each task over the STL method, which is formulated as $\Delta_{\mathrm{p}}=100\%\times\frac{1}{m}\sum_{i=1}^{m}\frac{1}{N_{i }}\sum_{j=1}^{N_{i}}\frac{(-1)^{s_{i, j}}(M_{i, j}-M_{i, j}^{\text{STL}})}{M_{i, j}^{\text{STL}}},$
where $N_i$ denotes the number of metrics for the $i$-th task, $M_{i,j}$ denotes the performance of an MTL method for the $j$-th metric in the $i$-th task, $M_{i, j}^{\text{STL}}$ is defined in the same way for the STL method, and $s_{i, j}$ is set to 0 if a larger value represents better performance for the $j$-th metric in $i$-th task and otherwise $s_{i, j}$ is set to 1.

\paragraph{Implementation Details.} 

For the Office-31 dataset, following RLW \cite{linreasonable}, we use a pre-trained ResNet-18 network as a shared backbone among tasks and a fully connected layer as the task-specific head. All input images are resized to $224\times 224$. The batch size is set to $64$. The cross-entropy loss is used for all tasks. 
The number of UL epochs $K$ is set to $100$. An Adam optimizer with the learning rate as $10^{-4}$ and the weight decay as $10^{-5}$ is used for updating model parameters $\omega$ in the UL subproblem.

\begin{table*}[!t]
\centering
\caption{Mean absolute error (MAE) on the QM9 dataset. Each experiment is repeated over 3 random seeds and the average is reported. 
The best results over baselines except STL are marked in \textbf{bold}. }
\vskip 0.1in
\resizebox{0.93\textwidth}{!}{
\begin{tabular}{lcccccccccccc}
\toprule
\multirow{1}{*}{\textbf{Methods}} & \multicolumn{1}{c}{$\bm{\mu}$} & \multicolumn{1}{c}{$\bm{\alpha}$} & \multicolumn{1}{c}{$\bm{\epsilon_{\mathrm{HOMO}}}$} & \multicolumn{1}{c}{$\bm{\epsilon_{\mathrm{LUMO}}}$} & \multicolumn{1}{c}{$\bm{\langle R^2 \rangle}$} & \multicolumn{1}{c}{\textbf{ZPVE}} & \multicolumn{1}{c}{$\bm{U_0}$} & \multicolumn{1}{c}{$\bm{U}$} & \multicolumn{1}{c}{$\bm{H}$} & \multicolumn{1}{c}{$\bm{G}$} & \multicolumn{1}{c}{$\bm{c_v}$} & \multirow{1}{*}{\bm{$\Delta_{\mathrm{p}}$}${\uparrow}$} \\
\midrule
STL & 0.062 & 0.192 & 58.82 & 51.95 & 0.529 & 4.52 & 63.69 & 60.83 & 68.33 & 60.31 & 0.069 & 0.00\\
\midrule
\multicolumn{13}{c}{\textit{multi-task learning methods}} \\
EW \cite{zhang2021survey} & 0.096 & 0.286 & \textbf{67.46} & 82.80 & 4.655 & 12.4 & 128.3 & 128.8 & 129.2 & 125.6 & 0.116 & -146.3\\
UW \cite{kgc18} & 0.336 & 0.382 & 155.1 & 144.3 & \textbf{0.965} & \textbf{4.58} & \textbf{61.41} & \textbf{61.79} & \textbf{61.83} & \textbf{61.40} & 0.116 & -92.35\\
MGDA \cite{sk18} & 0.181 & 0.325 & 118.6 & 92.45 & 2.411 & 5.55 & 103.7 & 104.2 & 104.4 & 103.7 & 0.110 & -103.0\\
PCGrad \cite{yu2020gradient} & 0.104 & 0.293 & 75.29 & 88.99 & 3.695 & 8.67 & 115.6 & 116.0 & 116.2 & 113.8 & 0.109 & -117.8\\
GradDrop \cite{chen2020just} & 0.114 & 0.349 & 75.94 & 94.62 & 5.315 & 15.8 & 155.2 & 156.1 & 156.6 & 151.9 & 0.136 & -191.4\\
GradVac \cite{wang2021gradient} & 0.100 & 0.299 & 68.94 & 84.14 & 4.833 & 12.5 & 127.3 & 127.8 & 128.1 & 124.7 & 0.117 & -150.7\\
CAGrad \cite{liu2021conflict} & 0.107 & 0.296 & 75.43 & 88.59 & 2.944 & 6.12 & 93.09 & 93.68 & 93.85 & 92.32 & 0.106 & -87.25\\

IMTL \cite{liu2021towards} &0.138 &0.344 &106.1& 102.9 &2.595 &7.84 &102.5& 103.0& 103.2& 100.8& 0.110 &-104.3\\
Nash-MTL \cite{navon2022multi} & 0.115 & \textbf{0.263} & 85.54 & 86.62 & 2.549 & 5.85 & 83.49 & 83.88 & 84.05 & 82.96 & 0.097 & -73.92\\
RLW \cite{linreasonable} & 0.112 & 0.331 & 74.59 & 90.48 & 6.015 & 15.6 & 156.0 & 156.8 & 157.3 & 151.6 & 0.133 & -200.9\\
\midrule
\multicolumn{13}{c}{\textit{first-order bi-level optimization methods}} \\
BVFIM \cite{liu2021value} & 0.107 & 0.325 & 73.18 & 98.97 & 5.336 & 21.4 & 200.1 & 201.2 & 201.8 & 195.5 & 0.148 & -228.5\\
BOME \cite{liubome} & 0.105 & 0.318 & 72.10 & 88.52 & 4.984 & 12.6 & 138.8 & 139.4 & 140.0 & 136.1 & 0.124 & -164.1\\
\midrule
\multicolumn{13}{c}{\textit{multi-objective bi-level optimization methods}} \\
MOML \cite{ye2021multi} & \textbf{0.083} & 0.347 & 74.87 & 80.57& 3.813 & 8.64 & 191.9 & 192.6 & 192.8 & 188.9 & 0.135 & -165.1\\
MoCo \cite{fernando2023mitigating} & 0.086 & 0.427 & 69.60 & \textbf{79.00} & 5.693 & 10.2 & 295.5 & 296.6 & 297.0 & 290.1 & 0.169 & -267.6\\
FORUM & 0.104 & 0.266 & 85.37 & 82.15 & 2.126 & 6.49 & 96.97 & 97.53 & 97.69 & 95.88 & \textbf{0.097} & \textbf{-73.36}\\
\bottomrule
\end{tabular}}
\vskip 0.1in
\label{tbl:qm9}
\end{table*}

For the NYUv2 dataset, following RLW \cite{linreasonable}, we use the DeepLabV3+ architecture \cite{ChenZPSA18}, which contains a ResNet-50 network with dilated convolutions as the shared encoder among all tasks and three Atrous Spatial Pyramid Pooling (ASPP) \cite{ChenZPSA18} modules as task-specific heads. All input images are resized to $288\times 384$. The batch size is set to $8$. The cross-entropy loss, $L_1$ loss, and cosine loss are used as the loss function of the three tasks, respectively. 
The total number of UL epochs $K$ is set to $200$. An Adam optimizer with the learning rate as $10^{-4}$ and the weight decay as $10^{-5}$ is used for updating model parameters $\omega$ in the UL subproblem. The learning rate of $\omega$ is halved after $100$ epochs.

For the QM9 dataset, following Nash-MTL \cite{navon2022multi}, we use a graph neural network \cite{gilmer2017neural} as the shared encoder, and a linear layer as the task-specific head. The targets of each task are normalized to have zero mean and unit standard deviation. The batch size is set to $128$. We use mean squared error (MSE) as the loss function. The total number of UL epochs $K$ is set to $300$. An Adam optimizer with a learning rate of $0.001$ is used for updating model parameters $\omega$ in the UL subproblem. A ReduceLROnPlateau scheduler \cite{paszke2019pytorch} is used to reduce the learning rate of $\omega$ once $\Delta_{\mathrm{p}}$ on the validation dataset stops improving. 

All methods are implemented based on the open-source \texttt{LibMTL} library \cite{lin2022libmtl}. For the proposed FORUM method, we set $\rho=0.1$, $\beta_k=(k+1)^{-\frac{3}{4}}$, use an SGD optimizer to update $T=5$ iterations in the LL subproblem and use an Adam optimizer to update the loss weight $\alpha$ in the UL subproblem. The UL step size $\mu$ is set to $0.001$ for all datasets, and the LL step size $\eta$ is set to $0.01$ for QM9 and $0.1$ for other datasets.
For the BOME, BVFIM, MOML, and MoCo methods, we use a similar configuration to the proposed FORUM method and perform a grid search for hyperparameters of each method. Specifically, we search LL learning rate $\eta$ over $\{0.05, 0.1, 0.5\}$ for both four methods, search $\rho$ over $\{0.1, 0.5, 0.9\}$ for BOME, search $\beta$ over $\{0.05, 0.1, 0.5, 1\}$ for BVFIM, and set $T=1$ for MOML and MoCo and $T=5$ for BOME and BVFIM.

\paragraph{Results.} 
Table \ref{tbl:mtl-office31} shows the results on Office-31. We can see that FORUM outperforms all baselines from different categories in terms of average classification accuracy and $\Delta_{\mathrm{p}}$, highlighting its effectiveness. The results on NYUv2 dataset are shown in Table \ref{tbl:mtl-nyu_dmtl}. As can be seen, only FORUM achieves better performance than STL in terms of  $\Delta_{\mathrm{p}}$. Moreover, FORUM performs well in the depth estimation and surface normal prediction tasks. The results on QM9 dataset are shown in Table \ref{tbl:qm9}. The QM9 dataset is a challenging dataset in MTL and none of the MTL methods performs better than STL, as observed in previous research \cite{navon2022multi}. We can see that FORUM again outperforms all the baselines in terms of $\Delta_{\mathrm{p}}$. Those results consistently demonstrate FORUM achieves state-of-the-art performance and is more effective than previous MOBLO methods such as MOML and MoCo.

\section{Conclusion}
In this paper, we propose FORUM, an efficient fully first-order gradient-based method for solving the multi-objective bi-level optimization problem. Specifically, we reformulate the original MOBLO problem to a constrained MOO problem and we propose a novel multi-gradient aggregation method to solve it. Compared with the existing MOBLO methods, FORUM does not require any hypergradient computation and thus is efficient. Theoretically, we provide a complexity analysis to show the efficiency of the proposed method and a non-asymptotic convergence guarantee for FORUM with a non-convex multi-objective UL subproblem. Moreover, empirical studies on different learning problems demonstrate the proposed FORUM method is effective and efficient. In particular, FORUM achieves state-of-the-art performance on three benchmark datasets under the setting of multi-task learning.

\section*{Acknowledgements}

This work is supported by National Key R\&D Program of China (No. 2022ZD0160300), NSFC key grant 62136005, NSFC general grant 62076118, and Shenzhen fundamental research program JCYJ20210324105000003.



\bibliography{mybibfile}

\clearpage
\onecolumn
\appendix

\section*{\centering {\Large Appendix }}

\section{Additional Materials}

\subsection{Gradient Approximation Error Bound}\label{sec:add1}

The following lemma shows that the gradient approximation error of $\nabla_z \widetilde{q}(z)$ exponentially decays w.r.t. the LL iterations $T$.

\begin{theorem_lemma}\label{app:lemma1}
Under Assumption \ref{assume:2} and suppose the step size $\eta$ satisfies $\eta\le \frac{2}{L_f+c}$, then we have $\|\nabla_z \widetilde{q}(z)- \nabla_{z} q(z)\|\le L_f (1-\frac{c\eta}{2})^T \|\tilde{\omega}^0 -\omega^*\|$.
\end{theorem_lemma}

\begin{proof}
According to Lemma 3 in \cite{sow2022constrained}, as $\eta\le \frac{2}{L_f+c}$, for a given $\alpha$, we have $\|\omega^T -\omega^*\|\le (1-\frac{c\eta}{2})^T \|\omega^0 -\omega^*\|$. Then for the approximated gradient $\nabla_{\alpha} f_{z}(\alpha,\tilde{\omega}^T)$, we have
\begin{align*}
        \|\nabla_z \widetilde{q}(z)- \nabla_{z} q(z)\| = \|\nabla_{\alpha}f(\alpha,\omega^*) - \nabla_{\alpha}f(\alpha,\tilde{\omega}^T)\|\le L_f\|\omega^*-\tilde{\omega}^T\|.
\end{align*}
Then we reach the conclusion.
\end{proof}

As $0<\eta\le \frac{2}{L_f+c}$, we have $0< \frac{c\eta}{2}< 1$. Therefore, the gradient approximation error exponentially decays w.r.t. the LL iterations $T$ according to Lemma \ref{app:lemma1}.

\subsection{Dual Problem of Problem (\ref{eq:modify_mgda})}\label{sec:add2}

The sub-problem can be rewritten equivalently as the following differentiable quadratic optimization
\begin{equation}\label{eq:tempaaaaa}
    d,\mu = \argmin_{d,\mu}(\frac{1}{2}\|d\|^2 +\mu)\ ~\mathrm{s.t.} \ ~  \langle \nabla \widetilde{q}(z_k),d\rangle \le -\phi,\ ~ \langle \nabla F_i(z_k),d\rangle\le \mu.
\end{equation}
Therefore we have 
\begin{equation}
    L = \frac{1}{2}\|d\|^2 + \mu + \sum_{i=1}^m \lambda_i(\langle \nabla F_i(z_k),d\rangle -\mu) + \nu (\langle \nabla \widetilde{q}(z_k),d\rangle+\phi), \ ~\mathrm{s.t.} \ ~  \sum_{i=1}^m \lambda_i =1.
\end{equation}
Differentiate with respect to $d$, and let $\nabla_d L =0$ we have
\begin{equation}
    d+ \sum_{i=1}^m \lambda\nabla F_i(z_k) + \nu\nabla \widetilde{q}(z_k) =0.
\end{equation}

Therefore, the gradient $d = -(\sum_{i=1}^m \lambda\nabla F_i(z_k) + \nu\nabla \widetilde{q}(z_k))$. Substitute it to  problem (\ref{eq:tempaaaaa}), we obtain that $\lambda$ and $\nu$ are the solution of 
\begin{equation}
    \min_{\lambda\in\Delta^{m-1},\nu\ge0} \frac{1}{2}\left\|\sum_{i=1}^m \lambda\nabla F_i(z_k) + \nu\nabla \widetilde{q}(z_k)\right\|^2 -\nu\phi. 
\end{equation}
For given $\lambda$, the above equation has a colsed form solution for $\nu$, 
\begin{equation}
    \nu(\lambda) = \max\left(\sum_{i=1}^m \lambda_i\pi_i(z), 0\right), \ ~\mathrm{s.t.} \ ~  \pi_i(z) = \frac{2\phi-\langle \nabla \widetilde{q}(z),\nabla F_i(z) \rangle}{\|\nabla \widetilde{q}(z)\|^2}.
\end{equation}
Therefore, the optimization problem of $\lambda$ becomes the following quadratic programming problem.
\begin{equation}\label{eq:FOMGDA}
 \lambda = \argmin_{\lambda\in\Delta^{m-1}}\left\|\sum_{i=1}^m\lambda_i\nabla F_i(z) + \nu(\lambda)\nabla \widetilde{q}(z)\right\|^2 - \nu(\lambda)\phi.
\end{equation}
which fits the result in Section \ref{sec:3.2}.

\subsection{Discussion on the Weights Sequence}\label{sec:add3}

For a standard non-convex MOO problem, i.e. $\min_z (F_1(z),F_2(z),\dots,F_m(z))$. The goal is to find a Pareto stationary point $z^*$, i.e., exists a $\lambda\in\Delta^{m-1}$ satisfies $\sum_{i=1}^m \lambda_i\nabla F_i(z^*)=0$. Suppose we use a gradient aggregated approach to update $z$, i.e., $z_{k+1}=z_k -\sum_{i=1}^m \mu\lambda_i^k\nabla F_i(z_k)$, where the gradient weights $\lambda$ can be obtained by any gradient-based MOO methods, such as MGDA \cite{desideri12} and CAGrad \cite{liu2021conflict}. We now show that the convergence of reaching a stationary point can only be guaranteed when $\{\lambda^k\}_{k=1}^K$ is a convergent sequence. 

Under Assumption \ref{assume:1}, since all objective functions are $L_F$-smooth, we have 
\begin{align*}
        \sum_{i=1}^m\lambda^{k}_i (F_i(z_{k+1}) -F_i(z_k)) &\le -\mu\| \sum_{i=1}^m\lambda^{k}_i \nabla F_i(z_{k})\|^2 + \frac{\sum_{i=1}^m\lambda^{k}_iL_F\mu^2}{2}\| \sum_{i=1}^m\lambda^{k}_i \nabla F_i(z_{k})\|^2. \\
\end{align*}
Let $\mu\le \frac{1}{L_F}$ and sum the above inequality over $k=0,1,\dots,K-1$ yields
\begin{align*}
    \sum_{k=1}^{K-1} \frac{\mu}{2}\| \sum_{i=1}^m\lambda^{k}_i \nabla F_i(z_{k})\|^2 \le \sum_{k=0}^{K-1} \sum_{i=1}^m\lambda^{k}_i (F_i(z_{k+1}) -F_i(z_k)).
\end{align*}
Therefore, we have
\begin{align*}
\frac{1}{K}\sum_{k=1}^{K-1}\| \sum_{i=1}^m\lambda^{k}_i \nabla F_i(z_{k})\|^2 \le \frac{\sum_{k=0}^{K-1}\sum_{i=1}^m(\lambda^{k}_i -\lambda^{k+1}_i)F(z_{k+1})}{\mu K}+ \frac{\sum_{i=1}^m (\lambda^{K-1}_iF(z_K)-\lambda^{0}_iF(z_0))}{\mu K}.
\end{align*}
Suppose the value of the functions $F_i$ are bounded, by carefully choosing the step size $\mu$, i.e., $\mu = \mathcal{O}(K^{-1/2})$, the second term of the right side of the above inequality can maintain a $K^{-1/2}$ convergent rate. However, if $\|\lambda^{k} -\lambda^{k+1}\|$ does not converge to zero, the first term is of order $\mathcal{O}(\mu^{-1})$. Thus it can not converge to zero for both constant step size or decreased step size. Therefore, the $\{\lambda^k\}_{k=1}^K$ should be a convergent sequence to guarantee the convergence of the gradient aggregated-based method for non-convex MOO problems. 

\section{Convergence Analysis}\label{sec:app_analysis}

\subsection{KKT Conditions}\label{sec:app_analysis_1}
Consider a general constrained MOO problem with one constraint such as problem (\ref{eq:FOMOBO_2}). The corresponding first-order Karush-Kuhn-Tucker (KKT) condition \cite{feng2018approximate} is said to hold at a feasible point $z^*\in\mathcal{Z}$ if there exist a vector $\lambda\in \Delta^{m-1}$ and $\nu\in\mathbb{R}_{+}$ such that the following three conditions hold,
\begin{align}\label{eq:Stationary_Condition}
     \sum_{i=1}^{m}\lambda_i \nabla F_i(z^*) + \nu \nabla q(z^*) =0, \ ~q(z^*) \le 0, \text{and} \ ~\nu q(z^*)=0.
\end{align}
Then $z^*$ is a local optimal point. The first condition is the stationarity condition, the second condition is the primal feasibility condition and the last condition is the complementary slackness condition. However, as we discussed in Section \ref{sec:3.1}, since the constraint function $q(z)$ is ill-posed, the complementary slackness condition can not be satisfied \cite{liubome,kwon2023fully}. To ensure our algorithm converges to a weak stationarity point, we measure the convergence by the stationarity condition and the feasibility condition.

\paragraph{Discussion on Pareto stationary.} The Pareto stationary is a first-order KKT stationary condition for the unconstrained MOO optimization problem. However, in this work, we reformulate MOBLO to an equivalent constrained MOO problem. Hence, the Pareto stationary cannot be used as a convergence criterion in our method. We measure the convergence by the local optimality condition of the constrained MOO problem, i.e., KKT stationary and feasibility conditions. 

\subsection{Lemmas}\label{sec:app_analysis_2}

We first provide the following lemmas for the LL subproblem under Assumption \ref{assume:2}. 
\begin{theorem_lemma}\label{LEMMA}
Under Assumption \ref{assume:2}, we have the following results. 
\begin{enumerate}[label=(\alph*)]

\item $\| \nabla\widetilde{q}(z)-\nabla q(z) \| \le L_f \|\tilde{\omega}^T -\omega^*(\alpha) \|$.

\item The function $\nabla q(z)$ is $L_q$-Lipschitz continuous w.r.t. $z$, where $L_q = L_f(2+\frac{L_f}{c})$.

\item If the step size of the LL subproblem satisfies $\eta \le \frac{2}{L_f} $, then for $\tilde{\omega}^0 = \omega$ and $\tilde{\omega}^{T+1} = \tilde{\omega}^{T}-\eta \nabla_{\omega}q(\alpha,\tilde{\omega}^{T})$, we have $q(\alpha,\tilde{\omega}^T)\le \Gamma(T) q(\alpha,\omega)$, where $\Gamma(T)$ represents an exponential decay w.r.t. $T$.

\item $\|\nabla_z q(z)\|^2 \le \frac{2L_q^2}{c}q(z)$

\end{enumerate}
\end{theorem_lemma}

\begin{proof}
\begin{enumerate}[label=(\alph*)]
\item We have $\| \nabla\widetilde{q}(z)-\nabla q(z) \| \le \| \nabla_{\alpha}f(\alpha,\tilde{\omega}^T) -\nabla_{\alpha}f(\alpha,\omega^*) \| \le L_f \|\tilde{\omega}^T -\omega^*(\alpha) \|$.

\item This result can be directly obtained from Lemma 5 in \cite{sow2022constrained}.

\item Since $\nabla_{\omega}q(z) =\nabla_{\omega}f(z) $ and $\nabla_{z}f(z)$ is $L_f$-Lipschitz continuous w.r.t $\omega$, $\nabla_{\omega}q(z)$ is $L_f$-Lipschitz continuous w.r.t $\omega$. Then we have 
\begin{align*}
    q(\alpha,\tilde{\omega}^{T+1}) &\le q(\alpha,\tilde{\omega}^{T})-(\eta-\frac{L_f\eta^2}{2})\|\nabla_{\omega} q(\alpha,\tilde{\omega}^{T})\|^2 \\ \nonumber
    & = q(\alpha,\tilde{\omega}^{T})-(\eta-\frac{L_f\eta^2}{2})\|\nabla_{\omega} f(\alpha,\tilde{\omega}^{T})\|^2\\ \nonumber
    &\le (1-c(2\eta-L_f\eta^2))  q(\alpha,\tilde{\omega}^{T}),
\end{align*}
where the second inequality is due to $ \|\nabla_{\omega}f(\alpha,\tilde{\omega}^{T})\|^2 \ge 2c(f(\alpha,\tilde{\omega}^{T})-f(\alpha,\omega^*)) =2cq(\alpha,\tilde{\omega}^{T})$.
Since $\tilde{\omega}^{0}=\omega$, we have $q(\alpha,\tilde{\omega}^{T}) \le (1-c(2\eta-L_f\eta^2))^T  q(\alpha,\omega)$. If $\eta \le \frac{2}{L_f}$, $2\eta-L_f\eta^2 \ge 0$. Let $\Gamma(T) = (1-c(2\eta-L_f\eta^2))^T$, which decays exponentially w.r.t. $T$. Then we reach the conclusion.

\item Since $\nabla q(\alpha,\omega^*(\alpha))=0$, we have $\|\nabla q(z)\|^2 = \|\nabla q(z) - \nabla q(\alpha,\omega^*(\alpha))\|^2\le L_q^2\|\omega-\omega^*(\alpha)\|^2$. Since $f(z)$ is $c$-strongly convex with respect to $\omega$, we have $\|\omega-\omega^*(\alpha)\|^2\le \frac{2}{c}(f(\alpha,\omega)-f(\alpha,\omega^*(\alpha))=\frac{2q(z)}{c}$, then we reach the conclusion.
\end{enumerate}
\end{proof}

Then for the Algorithm \ref{alg:example}, we have following result about the constraint function.
\begin{theorem_lemma}\label{LEMMA_q} 
    Under Assumption \ref{assume:2}, suppose the sequence $\{z_k\}_{k=0}^K$ generated by Algorithm \ref{alg:example} satisifes $q(z_k)\le B$, where $B$ is a positive constant. Then there exists a constant $C>0$, if $T\ge C$, the following results hold.
    \begin{enumerate}[label=(\alph*)]
        \item $q(z_k) \le \Gamma_1(k)B+ \mathcal{O}(\Gamma(T)+\mu)$, where $\Gamma_1(k)$ represents an exponential decay w.r.t $k$.
        \item $\|\nabla\widetilde{q}(z)-\nabla q(z)\| \le L_f\sqrt{\frac{\Gamma(T)}{c^2}}\|\nabla q(\alpha,\omega)\|$. 
        \item There exists a positive constant $C_b< 1$, such that $\|\nabla\widetilde{q}(z)\|\ge C_b\|\nabla q(z)\|$.
        \item $\sum_{k=0}^{K-1}\| \nabla \widetilde{q}(z_k)\|^2 = \mathcal{O}(\frac{1}{\mu}+ K\Gamma(T)+K\mu)$.
    \end{enumerate}

\end{theorem_lemma}
\begin{proof}
\begin{enumerate}[label=(\alph*)]
\item According to Lemma \ref{LEMMA} (d) and the boundedness assumption of $q(z_k)$, the gradient norm $\|\nabla q(z_k)\|$ is also bounded. Let $G(z_k)=\sum_{i=1}^m\tilde{\lambda}_i^k\nabla F_i(z_k)$, we have $d = -\mu(G(z_k)+\nu\nabla q(z_k))$ and $\nu = \max(\frac{\rho\|\nabla \widetilde{q}(z_k)\|^2-\langle \nabla \widetilde{q}(z_k),G(z_k) \rangle}{\|\nabla \widetilde{q}(z_k)\|^2},0)$. Then we can applying Lemma \ref{LEMMA} to Lemma 10 in \cite{liubome}, we obtain that there exist a constant $C>0$, if $T\ge C$, we have 
\begin{equation}
    q(\alpha_k,\omega_k) \le \Gamma_1(k)q(\alpha_0,\omega_0)+ \mathcal{O}(\Gamma(T)+\mu),
\end{equation}
where $\Gamma_1(k) = (1-\mu C_a)^k$ represents an exponential decay w.r.t $k$ and $C_a$ is a positive constant depending on $\eta$ and $c$. Then we reach the conclusion.

\item We have $\|\nabla \widetilde{q}(z) -\nabla q(z)\| \le L_f\|\tilde{\omega}^T -\omega^*(\alpha)\| \le L_f\sqrt{\frac{2(f(\alpha,\tilde{\omega}^T) - f(\alpha,\omega^*(\alpha)))}{c}}$, where the first inequality is due to Lemma \ref{LEMMA} (a). Then we have 
\begin{align*}
    \| \nabla \widetilde{q}(z)-\nabla q(z)\|
    \le L_f\sqrt{\frac{2\Gamma(T)q(z)}{c}} 
    \le L_f\sqrt{\frac{\Gamma(T)}{c^2}}\|\nabla q(\alpha,\omega)\|,
\end{align*}
where the first inequality is due to Lemma \ref{LEMMA} (c) and the second inequality is due to the strongly convex assumption of $f(\alpha,\omega)$ w.r.t $\omega$ and $\|\nabla_{\omega}q(\alpha,\omega)\|\le \|\nabla q(z)\|$. 

\item By the triangle inequality and Lemma \ref{LEMMA_q} (b), we obtain
\begin{align*}
    \| \nabla \widetilde{q}(z)\|  \ge  \|\nabla q(z)\| - \|\nabla \widetilde{q}(z)-\nabla q(z)\| \ge (1- L_f\sqrt{\frac{\Gamma(T)}{c^2}})\|\nabla q(z)\|. 
\end{align*}
Then there exists a positive constant $C>0$ such that when $T\ge C$, we have $0<  L_f\sqrt{\frac{\Gamma(C)}{c^2}}< 1$. Let $C_b= 1-L_f\sqrt{\frac{\Gamma(C)}{c^2}}$, then we have $\| \nabla \widetilde{q}(z)\| \ge C_b \|\nabla q(z)\|$ and $C_b< 1$.

\item By the triangle inequality and Lemma \ref{LEMMA_q} (b), we obtain
\begin{align*}
    \| \nabla \widetilde{q}(z)\|  \le \| \nabla \widetilde{q}(z)-\nabla q(z)\| + \|\nabla q(z)\| \le (1+ L_f\sqrt{\frac{\Gamma(T)}{c^2}})\|\nabla q(z)\|.
\end{align*}
By defining $C_e = \left(1+ L_f\sqrt{\frac{\Gamma(T)}{c^2}}\right)^2$, we have  
\begin{align*}
    \sum_{k=0}^{K-1}\| \nabla \widetilde{q}(z_k)\|^2&\le \sum_{k=0}^{K-1} C_e \|\nabla q(z_k)\|^2 \le \sum_{k=0}^{K-1} \frac{2L_q^2C_e}{c}(\Gamma_1(k)B+ \mathcal{O}(\Gamma(T)+\mu)),
\end{align*}
where the second inequality is due to Lemma \ref{LEMMA} (d) and Lemma \ref{LEMMA_q} (a). Since $\sum_{k=0}^{K}(1-\mu C_a)^k = \mathcal{O}(\frac{1}{\mu})$ and $C_e$ decays to $1$ as $T\to +\infty$, we have $\sum_{k=0}^{K-1}\| \nabla \widetilde{q}(z_k)\|^2 = \mathcal{O}(\frac{1}{\mu}+ K\Gamma(T)+K\mu)$.
\end{enumerate}
\end{proof}

\begin{remark}
Lemma \ref{LEMMA_q} (a) uses the result of Lemma 10 in \citep{liubome}. It is worth noting that the assumptions 1 and 3 in \citep{liubome}, which assume that the LL subproblem both satisfies PL-inequality and the gradient boundedness, are self-contradicted. In our analysis, we address this theoretical problem by assuming that $q(z_k)$ is bounded for the generated sequence $\{z_k\}_{k=1}^{K}$. This means the convergence can still be held locally for a finite sequence $\{z_k\}_{k=1}^{K}$.
\end{remark}

\subsection{Proof of the Theorem \ref{thm:2}}\label{sec:app_analysis_3}

Since $F_i(z)$ is $L_F$-Lipschitz continuous, we have
\begin{align*}
        \sum_{i=1}^m\tilde{\lambda}^{k}_i (F_i(z_{k+1}) -F_i(z_k)) &\le \mu \left\langle \sum_{i=1}^m\tilde{\lambda}^{k}_i \nabla F_i(z_{k}),d_k \right\rangle + \frac{\sum_{i=1}^m\tilde{\lambda}^{k}_iL_F\mu^2}{2}\|d_k\|^2 \\
        &= \mu \left\langle -\nu_k\nabla \widetilde{q}(z_k)-d_k,d_k \right\rangle + \frac{L_F\mu^2}{2}\|d_k\|^2 \\
        &= (\frac{L_F\mu^2}{2}-\mu)\|d_k\|^2- \mu\nu_k\langle \nabla \widetilde{q}(z_k),d_k \rangle.
\end{align*}
According to the complementary slackness condition of problem (\ref{eq:modify_mgda}), We have $\nu_k(\langle \nabla \widetilde{q}(z_k),d_k \rangle + \phi_k) = 0$, where $\nu_k=\nu(\tilde{\lambda}^k)$. Therefore $-\nu_k\langle \nabla \widetilde{q}(z_k),d_k \rangle = \nu_k\frac{\rho}{2}\|\nabla \widetilde{q}(z_k)\|^2$. Then if $\mu\le \frac{1}{L_F}$, we have 
\begin{equation}\label{eq:proof_1}
    \sum_{i=1}^m\tilde{\lambda}^{k}_i (F_i(z_{k+1}) -F_i(z_k))  \le -\frac{\mu}{2}\|d_k\|^2+ \frac{\rho\mu\nu_k}{2}\|\nabla \widetilde{q}(z_k)\|^2. 
\end{equation}
Let $G(z_k)=\sum_{i=1}^m\tilde{\lambda}_i^k\nabla F_i(z_k)$, we have  $\nu_k\|\nabla \widetilde{q}(z_k)\|^2 \le \rho\|\nabla \widetilde{q}(z_k)\|^2 - \langle G(z_k),\nabla \widetilde{q}(z_k)\rangle$.
Summing the inequality (\ref{eq:proof_1}) over $k=0,1,\dots,K-1$ yields
\begin{align*}
    \sum_{k=0}^{K-1} \sum_{i=1}^m\tilde{\lambda}^{k}_i (F_i(z_{k+1}) -F_i(z_k))  & \le \sum_{k=0}^{K-1}( -\frac{\mu}{2}\|d_k\|^2+ \frac{\rho\mu\nu_k}{2}\|\nabla \widetilde{q}(z_k)\|^2) \\
    & \le-\frac{\mu}{2}\sum_{k=0}^{K-1}\|d_k\|^2 + \sum_{k=0}^{K-1}\frac{\rho^2\mu}{2}\|\nabla \widetilde{q}(z_k)\|^2 - \sum_{k=0}^{K-1}\frac{\rho\mu}{2}\langle G(z_k),\nabla \widetilde{q}(z_k)\rangle \\
    & \le -\frac{\mu}{2}\sum_{k=0}^{K-1}\|d_k\|^2 + \sum_{k=0}^{K-1}\frac{\rho^2\mu}{2}\|\nabla \widetilde{q}(z_k)\|^2 + \sum_{k=0}^{K-1}\frac{\rho\mu M}{2}\|\nabla \widetilde{q}(z_k)\|,
\end{align*}
where the last inequality is by Cauchy-Schwarz inequality and $\|G(z_k)\|\le \|\nabla_z F_i(z_k)\| \le M$. Therefore, we further have 
\begin{equation}\label{eq:proof_2}
    \sum_{k=0}^{K-1}\|d_k\|^2 \le \frac{2\sum_{k=0}^{K-1} \sum_{i=1}^m\tilde{\lambda}^{k}_i (F_i(z_{k+1}) -F_i(z_k))}{\mu} + \sum_{k=0}^{K-1}\rho^2\|\nabla \widetilde{q}(z_k)\|^2 + \sum_{k=0}^{K-1}\rho M\|\nabla \widetilde{q}(z_k)\|.
\end{equation}
For the first term of the right-hand side of the inequality (\ref{eq:proof_2}), we have 
\begin{align*}
\sum_{k=0}^{K-1} \sum_{i=1}^m\tilde{\lambda}^{k}_i (F_i(z_{k+1}) -F_i(z_k))&= \sum_{k=0}^{K-1}\sum_{i=1}^m(\tilde{\lambda}^{k}_i -\tilde{\lambda}^{k+1}_i)F(z_{k+1})+ \sum_{i=1}^m (\tilde{\lambda}^{K-1}_iF(z_K)-\tilde{\lambda}^{0}_iF(z_0)) \\
&\le \sum_{k=0}^{K-1}\sum_{i=1}^m|\tilde{\lambda}^{k}_i -(1-\beta) \tilde{\lambda}^{k}_i- \beta\lambda^{k+1}_i|M+ 2M \\
&\le \sum_{k=0}^{K-1}\beta\sum_{i=1}^m|\tilde{\lambda}^{k}_i -\lambda^{k+1}_i|M+ 2M.
\end{align*}
Since $\lambda\in\Delta^{m-1}$, we have $|\tilde{\lambda}^{k}_i -\lambda^{k+1}_i|\le 2$. Then we have 
\begin{align*}
    \sum_{k=0}^{K-1}\|d_k\|^2 & \le \frac{2mMK\beta+2M}{\mu}+ \sum_{k=0}^{K-1}\rho^2\|\nabla \widetilde{q}(z_k)\|^2 + \rho M\sqrt{K}\sqrt{\sum_{k=0}^{K-1}\|\nabla \widetilde{q}(z_k)\|^2} \\
    & = \mathcal{O}(\frac{K\beta}{\mu})+ \mathcal{O}(\frac{M}{\mu})+ \mathcal{O}(\frac{1}{\mu}+ K\Gamma(T)+K\mu) + \mathcal{O}(\sqrt{\frac{K}{\mu}}+ K\sqrt{\Gamma(T)}+K\sqrt{\mu}) \\
    & = \mathcal{O}(\frac{K\beta}{\mu} + K\Gamma(T) + \sqrt{\frac{K}{\mu}}+K\sqrt{\mu}),
\end{align*}
where the first inequality is by Holder's inequality and the first equality is due to Lemma \ref{LEMMA_q} (d). For the measure of stationarity $\mathcal{K}(z_k)$, we obtain that
\begin{equation}\label{eq:proof_3}
    \mathcal{K}(z_k) = \|G(z_k)+\nu_k\nabla q(z_k)\|^2 \le 2\|d_k\|^2+ 2\|\nu_k(\nabla\widetilde{q}(z_k)-\nabla q(z_k))\|^2. 
\end{equation}

According to \ref{LEMMA} (d), we have $\|\nabla q(z_k)\| \le \sqrt{\frac{2BL_q^2}{c}}$. Then we obtain
\begin{align*}
    \|\nu_k(\nabla\widetilde{q}(z_k)-\nabla q(z_k))\| &\le \left|\rho - \frac{\langle G(z_k),\nabla \widetilde{q}(z_k)\rangle}{\|\nabla \widetilde{q}(z_k)\|^2}\right|\|\nabla\widetilde{q}(z_k)-\nabla q(z_k)\| \\
    &\le \rho\|\nabla\widetilde{q}(z_k)-\nabla q(z_k)\| + |\langle G(z_k),\frac{\nabla \widetilde{q}(z_k)}{\|\nabla \widetilde{q}(z_k)\|}\rangle |\frac{\|\nabla\widetilde{q}(z_k)-\nabla q(z_k)\|}{\|\nabla\widetilde{q}(z_k)\|} \\
    & \le L_f\sqrt{\frac{\Gamma(T)}{c^2}}(\rho\|\nabla q(z_k)\| + \frac{1}{C_b}|\langle G(z_k),\frac{\nabla \widetilde{q}(z_k)}{\|\nabla \widetilde{q}(z_k)\|}\rangle|) \\
    &\le L_f\sqrt{\frac{\Gamma(T)}{c^2}}(\rho L_q\sqrt{\frac{2B}{c}}+ \frac{M}{C_b}),
\end{align*}
where the third inequality is due to Lemma \ref{LEMMA_q} (b) and (c), and the last inequality is due to the Cauchy-Schwarz inequality. Therefore, we have $\|\nu_k(\nabla\widetilde{q}(z_k)-\nabla q(z_k))\|^2= \mathcal{O}(\Gamma(T))$. Then we can get 
\begin{align*}
    \min_{k<K}\mathcal{K}(z_k) & \le \frac{1}{K}\sum_{k=0}^{K-1}\mathcal{K}(z_k)    \\& = 2 \mathcal{O}({\frac{K\beta}{\mu K} + \Gamma(T) + \sqrt{\frac{1}{\mu K}}+\sqrt{\mu}}) + 2 \mathcal{O}(\Gamma(T)) \\
    & = \mathcal{O}({\frac{\beta}{\mu} + \Gamma(T) + \sqrt{\frac{1}{\mu K}}+\sqrt{\mu}}),
\end{align*}
where the first equality is due to Eq. (\ref{eq:proof_3}). According to Lemma \ref{LEMMA_q} (a), we obtain  $q(z_k) = \mathcal{O}(\Gamma_1(k)+ \Gamma(T)+\mu)$. Thus we get
\begin{equation*}
    \max\left\{\min_{k<K}\mathcal{K}(z_k),q(z_k)\right\} = 
       \mathcal{O}\left(\sqrt{\mu}+ \sqrt{\frac{1}{\mu K}}+ \frac{\beta}{\mu} +\Gamma(T)\right).
\end{equation*}
Let $\mu=\mathcal{O}(K^{-1/2})$ and $\beta=\mathcal{O}(K^{-3/4})$, we reach the conclusion.

\section{Synthetic MOBLO} \label{sec:synthetic_example}

In this section, we use one synthetic MOBLO problem to illustrate the convergence of the proposed FORUM method. We first consider the following problem,
\begin{align}\label{eq:example1}
    &\min_{\alpha\in \mathbb{R},\omega\in \mathbb{R}^2}(\|\omega - (1,\alpha)\|^2_2 , \|\omega - (2,\alpha)\|^2_2) \ ~ \quad\text{s.t.\ ~} \omega\in \argminl_{\omega\in \mathbb{R}^2}\|\omega-\alpha\|_2^2,
\end{align}
where $(\cdot,\cdot)$ denotes a two-dimensional vector and $\omega = (\omega_1,\omega_2)$. 
Problem (\ref{eq:example1}) satisfies all the assumptions required in Section \ref{sec:con_analysis}. By simple calculation, we can find that the optimal solution set of problem (\ref{eq:example1}) is $\mathcal{P}=\{(\alpha,\omega) \mid \alpha = \omega_1 = \omega_2 =c, c\in [1,2]\}$. 

We apply GD optimizer for both UL and LL subproblems and the step sizes are set to $\mu = 0.3$ and $\eta =0.05$ for all methods.  We run $50$ LL iterations to ensure that for a given $\alpha$, they all reach the minimum point for the LL subproblem. For FORUM, we set $\rho=0.3$ and $\beta_k = (k+1)^{-3/4}$. The result is evaluated by calculating the Euclidean distance between solution $z$ and the optimal solution set $\mathcal{P}$, which is denoted by $\mathcal{E}=\text{dist}(z,\mathcal{P})$.

\begin{figure*}[!t]
\centering
\subfigure[$\mathcal{E}$ vs. $K$.]{\label{fig:example1_a}\includegraphics[width=0.24\textwidth]{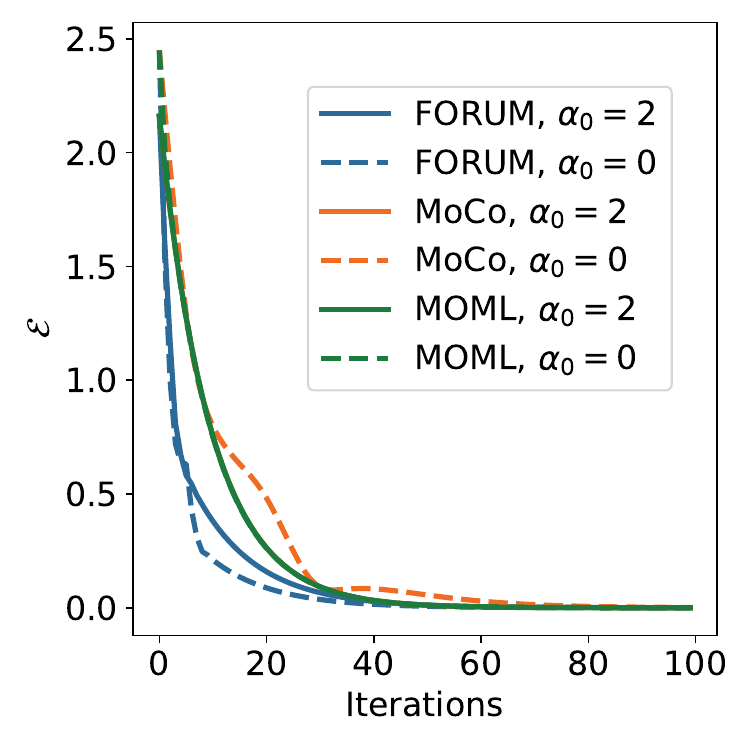}}
\subfigure[$\mathcal{E}$ vs. $K$.]
{\label{fig:example1_b}\includegraphics[width=0.24\textwidth]{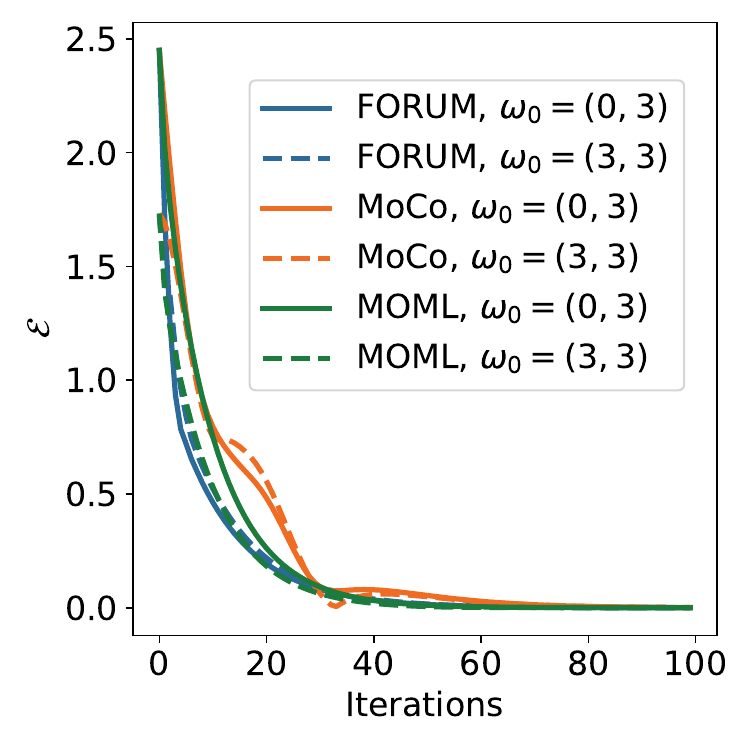}}
\subfigure[$\mathcal{K}(z)$ vs. $K$.]
{\label{fig:example1_c}\includegraphics[width=0.24\textwidth]{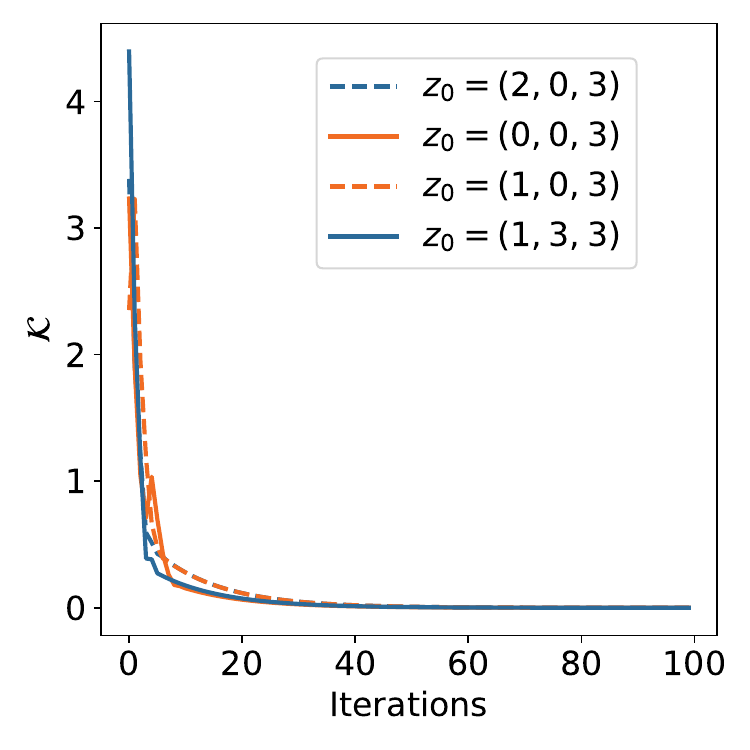}}
\subfigure[$q(z)$ vs. $K$.]
{\label{fig:example1_d}\includegraphics[width=0.24\textwidth]{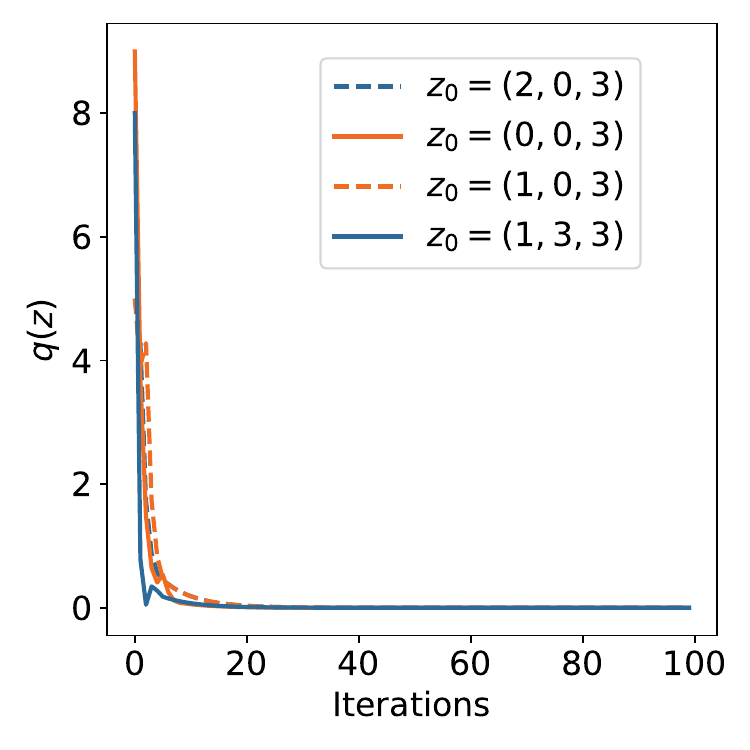}}
\vskip 0.05in
\caption{Results on the problem (\ref{eq:example1}) with different initialization points. \textbf{(a)}: Fix $\omega_0=(0,3)$ and vary $\alpha_0 = 0, 2$. The optimality gap $\mathcal{E}$ curves. \textbf{(b)}: Fix $\alpha_0 = 2$ and vary $\omega_0=(0,3), (3,3)$. The optimality gap $\mathcal{E}$ curves. \textbf{(c)}: The stationarity gap $\mathcal{K}(z)$ curves. \textbf{(d)}: The value of the constraint function $q(z)$ curves.}
\vskip 0.15in
\label{fig:example1}
\end{figure*}
Figures \ref{fig:example1_a} and \ref{fig:example1_b} show the numerical results of the MOML, MoCo, and FORUM methods with different initializations. It can be observed that the proposed FORUM method can achieve an optimal solution in all the settings, i.e., $\mathcal{E}\to 0$, and different initializations only slightly affect the convergence speed. 
Figures \ref{fig:example1_c} and \ref{fig:example1_d} show that both $\mathcal{K}(z)$ and $q(z)$ converge to zero in all the settings. Thus FORUM solves the corresponding constrained optimization problem. This result demonstrates our convergence result in Section \ref{sec:con_analysis}.

\section{Comparison with Non-Gradient-Based Methods}

We compare the proposed FORUM method with two non-gradient-based methods, i.e., the Bayesian method and the NSGA-II method, on the multi-objective data hyper-cleaning problem. The experimental setting is the same as in Section 5.1. For these two non-gradient-based methods, in each iteration, we first learn a model $\omega$ by SGD with the given $\alpha$, then compute the upper-level objectives, and finally update $\alpha$ by the non-gradient-based method. We implement the Bayesian method with the open-source \texttt{Hyperopt} library\footnote{\url{https://github.com/hyperopt/hyperopt}}. The NSGA-II method is implemented based on the open-source \texttt{Optuna} library\footnote{\url{https://github.com/optuna/optuna}}. The experimental results are shown in Table \ref{tbl:non-gradient}. The results show that the proposed FORUM is more effective than those non-gradient-based methods.

\begin{table}[!h]
\centering
\caption{Comparison with non-gradient-based methods. Each experiment is repeated over $3$ random seeds, and the mean as well as the standard deviation is reported. The best result is marked in \textbf{bold}.} 
\vskip 0.05in
\begin{tabular}{lcccc}
\toprule
\multirow{2.5}{*}{\textbf{Methods}} & \multicolumn{2}{c}{\textbf{MNIST}} & \multicolumn{2}{c}{\textbf{FashionMNIST}}\\
\cmidrule(lr){2-3} \cmidrule(l){4-5}
& \textbf{Accuracy (\%)} & \textbf{F1 Score} & \textbf{Accuracy (\%)} & \textbf{F1 Score}\\
\midrule
Bayesian method & $85.70_{\pm0.92}$ & $85.71_{\pm0.93}$ & $76.36_{\pm1.93}$ & $76.17_{\pm2.10}$\\
NSGA-II & $85.79_{\pm1.97}$ & $85.76_{\pm1.96}$ & $76.39_{\pm1.95}$ & $75.78_{\pm2.02}$\\
FORUM ($T=16$) (\textbf{ours}) & $\textbf{90.79}_{\pm0.33}$ & $\textbf{90.79}_{\pm0.33}$ & $\textbf{82.37}_{\pm1.00}$ & $\textbf{82.10}_{\pm1.16}$\\
\bottomrule
\end{tabular}
\vskip 0.1in
\label{tbl:non-gradient}
\end{table}

\end{document}